\documentclass[10pt,twocolumn,letterpaper]{article}

\usepackage[pagenumbers]{cvpr} 

\usepackage{graphicx}
\usepackage{amsmath}
\usepackage{amssymb}
\usepackage{booktabs}

\usepackage{xcolor, colortbl}
\definecolor{citecolor}{HTML}{0071bc}
\definecolor{myblue}{rgb}{0.88,0.98,0.98}
\definecolor{mygreen}{rgb}{0, 0.5, 0.1}
\definecolor{myred}{rgb}{0.84, 0.04, 0.33}
\definecolor{mygray}{gray}{0.92}
\usepackage[pagebackref=false,breaklinks=true,letterpaper=true,colorlinks,citecolor=citecolor,bookmarks=false]{hyperref}

\usepackage{caption}
\usepackage{subcaption}
\usepackage{float}
\usepackage{bbm}
\usepackage{booktabs}
\usepackage{pifont}
\newcommand{\cmark}{\ding{51}}%
\newcommand{\xmark}{\ding{55}}%
\usepackage{multirow}
\usepackage{multicol}
\usepackage{threeparttable}

\usepackage[capitalize]{cleveref}
\crefname{section}{Sec.}{Secs.}
\Crefname{section}{Section}{Sections}
\Crefname{table}{Table}{Tables}
\crefname{table}{Tab.}{Tabs.}

\def\cvprPaperID{0001} 
\def\confName{CVPR}
\def\confYear{2022}

\begin{document}

\title{Voxel Field Fusion for 3D Object Detection }

\author{Yanwei Li$^{1}$\quad Xiaojuan Qi$^{2}$\quad Yukang Chen$^{1}$\quad Liwei Wang$^{1}$\quad \\
Zeming Li$^{3}$\quad Jian Sun$^{3}$\quad Jiaya Jia$^{1,4}$  \\[0.2cm]
The Chinese University of Hong Kong$^{1}$\quad The University of Hong Kong$^{2}$ \\
MEGVII Technology$^{3}$\quad SmartMore$^{4}$
}

\maketitle


\begin{abstract}
   In this work, we present a conceptually simple yet effective framework for cross-modality 3D object detection, named voxel field fusion.
   The proposed approach aims to maintain cross-modality consistency by representing and fusing augmented image features as a ray in the voxel field.
   To this end, the learnable sampler is first designed to sample vital features from the image plane that are projected to the voxel grid in a point-to-ray manner, which maintains the consistency in feature representation with spatial context.
   In addition, ray-wise fusion is conducted to fuse features with the supplemental context in the constructed voxel field.
   We further develop mixed augmentor to align feature-variant transformations, which bridges the modality gap in data augmentation.
   The proposed framework is demonstrated to achieve consistent gains in various benchmarks and outperforms previous fusion-based methods on KITTI and nuScenes datasets.
   Code is made available at \href{https://github.com/dvlab-research/VFF}{https://github.com/dvlab-research/VFF}.\footnote{Part of the work was done in MEGVII Research.}
\end{abstract}

\section{Introduction}
\label{sec:intro}
Object detection in 3D scenes is regarded as a vital task to provide accurate perception for real-world applications.
Over the past decades, research attentions~\cite{zhou2018voxelnet,yan2018second,yang2018pixor,lang2019pointpillars,shi2020pvrcnn} have been dedicated to 3D object detection from raw point clouds.
Due to the inherent properties of LiDAR sensors, the captured point clouds are usually sparse and cannot provide sufficient context to distinguish among hard cases in distant or occluded regions, which consequently yields inferior performance in such scenarios.
However, in safety-critical applications like autonomous driving, the frequently occurred miss-detection is unacceptable.

\begin{figure}[t!] 
\centering
\begin{subfigure}[t]{0.48\linewidth}
\centering
\includegraphics[width=\linewidth]{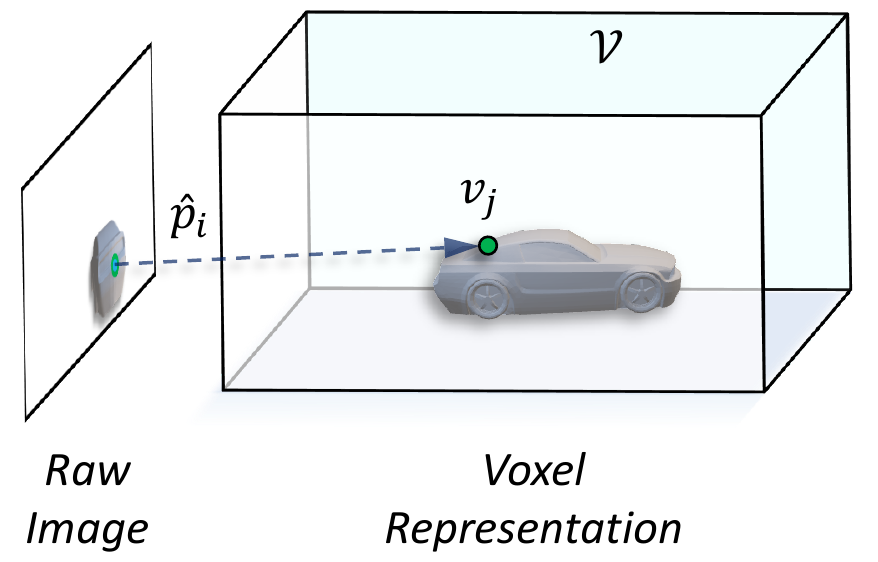}
\caption{Point-to-point manner}
\label{fig:intro_raw}
\end{subfigure}
\hfill
\begin{subfigure}[t]{0.48\linewidth}
\centering
\includegraphics[width=\linewidth]{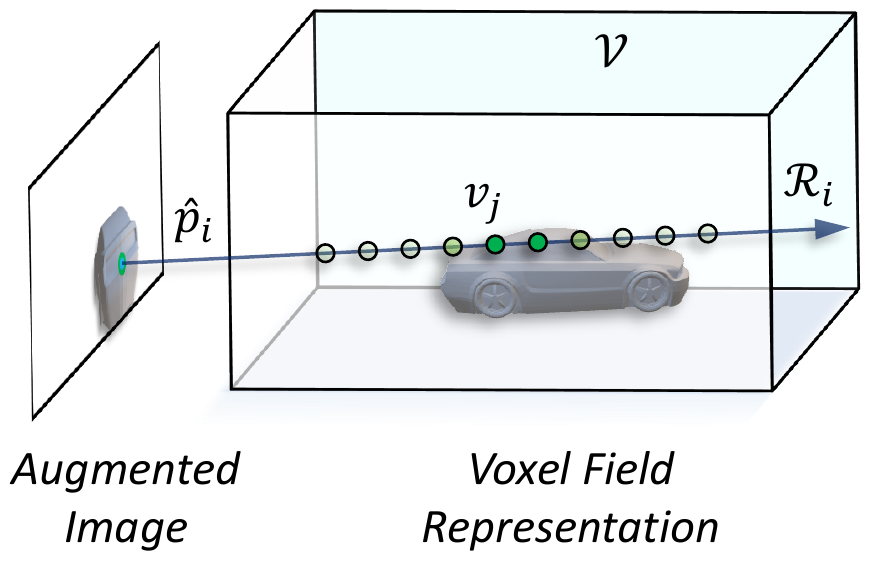}
\caption{Point-to-ray manner}
\label{fig:intro_ours}
\end{subfigure}
\caption{Compared with previous work~\cite{liang2018continuous,zhang2020moca} in~\ref{fig:intro_raw} that projects feature from raw image to voxel and represents in a {\em point-to-point} manner, the proposed method in~\ref{fig:intro_ours} projects feature from augmented image to voxel field and represents in a {\em point-to-ray} manner. Dotted and solid arrow denote point- and ray-level projection.}
\label{fig:intro}
\end{figure}

To address this issue, previous studies introduce image features in the cross-modality fusion~\cite{sindagi2019mvx,huang2020epnet,vora2020pointpainting}.
The main challenge is to maintain the cross-modality consistency in this process that might be damaged in {\em feature representation} considering context deficiency and density variance, and {\em data augmentation} for cross-modality misalignment.
In particular, previous work represents image features in a {\em point-to-point} manner in Figure~\ref{fig:intro_raw}, which conducts fusion in each single point, constrained by the sparsity of point clouds.
In this case, the rich context cues from images cannot be well utilized because adjacency in the image plane cannot be guaranteed in 3D space.
Meanwhile, given augmented point clouds, traditional approaches~\cite{liang2018continuous,zhang2020moca} usually keep raw images  unchanged and reverse transformations in point clouds for pairwise correspondence.
However, because of the flipping and scaling variance in 2D convolutions, the asynchronous augmentation brings cross-modality misalignment and instability.

In this paper, we propose a new cross-modality framework, called {\em voxel field fusion} (VFF).
Mixed augmentation for both modalities is first applied for data-level pre-processing.
As briefly illustrated in Figure~\ref{fig:intro_ours}, VFF projects augmented image features to the voxel grid and represents it in a point-to-ray manner, called voxel field similar to that in neural rendering~\cite{mildenhall2020nerf,liu2020neural}.
In this way, representations of both modalities are well aligned, and surrounding spatial context is replenished in voxel field.
In short, the key idea of VFF is to maintain modality consistency by representing and fusing augmented image feature as a ray in voxel field.

The image-to-voxel dense rendering is usually resource-intensive or requires extra models for depth prediction~\cite{reading2021caddn,you2020pseudo++}.
To facilitate this process, we draw inspiration from recent advances in neural rendering~\cite{mildenhall2020nerf,liu2020neural,yu2021pixelnerf} and propose a learnable sampler and ray-wise fusion for efficient ray construction and cross-modality fusion.
In particular, instead of random sampling~\cite{mildenhall2020nerf}, learnable sampler is designed to select image features for interaction within the activated area with high responses, where features are represented in a point-to-ray manner as aforementioned.
Then, ray-wise fusion is conducted in the voxel field according to the predicted score of each voxel along the ray.
For the misalignment in augmentation, the mixed augmentor is further proposed to bridge this gap by aligning feature-variant augmentation (flipping and scaling) in image level.

With the above designs, the cross-modality consistency can be maintained from the aspect of feature representation and data augmentation in an end-to-end manner. 
Generally, the proposed VFF is distinguished from two aspects.
First, it projects image features in a {\em point-to-ray} manner and represents, as well as fuses them, in the voxel field, which eliminates the modality gap and provides accurate 3D context to detect hard cases.
Second, it efficiently samples high-responded features from augmented images, which enables the network to construct each ray on the fly.

The overall framework, called {\em voxel field fusion}, can be easily instantiated with various voxel-based backbones for 3D object detection, which is fully elaborated in Section~\ref{sec:method}.
Extensive empirical studies of the designed workflow are conducted in Section~\ref{sec:experiment} to reveal the effect of each component.
We further report experimental results on two widely-adopted datasets, namely KITTI~\cite{geiger2012kitti} and nuScenes~\cite{caesar2020nuscenes}.
The proposed VFF is proved to achieve consistent increases over various benchmarks and attains significant gain with {\bf 2.2}\% AP over strong baselines on {\em hard} cases of KITTI {\em test} set.
Meanwhile, it surpasses previous fusion-based methods by a large margin and achieves leading performance on nuScenes {\em test} set with {\bf 68.4}\% mAP and {\bf 72.4}\% NDS.

\section{Related Work}
\label{sec:related}

\noindent
\textbf{LiDAR-based 3D Detection.}
Given point clouds as input, traditional LiDAR-based approaches are usually distinguished by their representation for irregular data, {\em e.g.,} grid and point.
Grid-based methods project the point clouds to regular grids and process them by 2D or 3D networks. 
The methods with 2D networks usually construct 2D bird-view grids~\cite{chen2017mv3d,yang2018pixor,yang2018hdnet} or pseudo image~\cite{lang2019pointpillars} and generating 3D bonding boxes on top of it. 
Meanwhile, the approaches with 3D networks construct 3D voxels from the divided point clouds and predict boxes with detection heads~\cite{yan2018second,deng2020voxelrcnn,yin2021centerpoint}. Point-based methods directly handle raw point cloud with set abstraction~\cite{qi2017pointnet++} and generate 3D proposals on top of it~\cite{shi2019pointrcnn,yang2019std,qi2019votenet,qi2020imvotenet}.
Given the sparse property of point cloud, the recognition ability is limited due to lack of texture features, especially in real scenes with multiple categories like nuScenes~\cite{caesar2020nuscenes} dataset.

\vspace{0.5em}
\noindent
\textbf{Image-based 3D Detection.}
Previous image-based methods construct the network and extract features from pure monocular or multiple images for 3D box prediction.
Given a single image, several monocular-based approaches~\cite{brazil2019m3d,simonelli2019disentangling,wang2021fcos3d} try to regress and predict 3D boxes directly, while others propose to construct middle-level representation and perform detection on top of it~\cite{wang2019pseudo,you2020pseudo++}.
Because of the depth requirement in 3D detection, previous work also tries to enhance the ability from depth estimation~\cite{chen20153d,shi2020distance,reading2021caddn}.
Another stream for relatively accurate depth is utilizing stereo or multi-view images to construct 3D geometry volume~\cite{yao2018mvsnet,chen2019pointmvsnet,chen2020dsgn} and conduct object detection on top of it.
Although depth estimated from multi-view is much better than that from a single image, it still lags behind the accurate point cloud from LiDAR.

\begin{figure*}[th!]
\centering
\includegraphics[width=\linewidth]{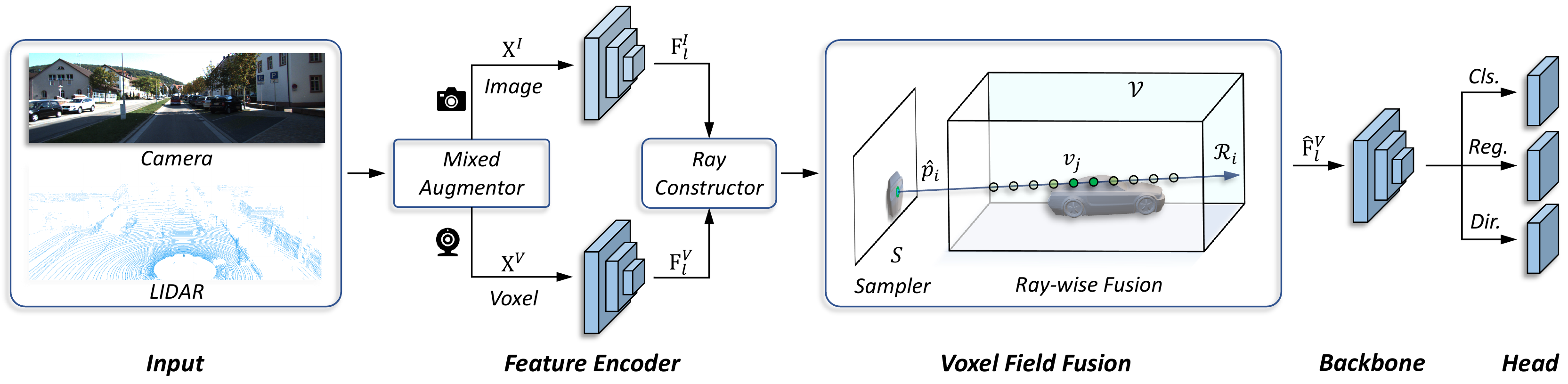} 
\caption{
The framework for 3D object detection with {\em voxel field fusion}. 
In particular, inputs with different modalities are firstly processed with the mixed augmentor that is adopted for training only.
Then, features of both modalities are respectively extracted in feature encoder, where the correspondence is established in ray constructor.
In voxel field fusion, the vital image feature for interaction is selected with the designed sampler.
And the ray-wise fusion is then conducted with the high-responded feature along each ray.
With the fused and newly generated feature in the voxel field, the following detection backbone and head are applied to predict the final 3D proposals.
}
\label{fig:main}
\end{figure*}

\vspace{0.5em}
\noindent
\textbf{Cross-modality Fusion.}
With inherent limitation of every single modality, there are several methods to combine the strength of image and LiDAR with cross-modality fusion. 
In particular, point- and proposal-level fusion are introduced to combine features from different modalities. Point-level fusion~\cite{liang2018continuous,huang2020epnet,vora2020pointpainting} is usually applied in early stage of the network, while the proposal-based manner~\cite{chen2017mv3d,ku2018avod,yoo20203dcvf} is often adopted in the late stage for instance-level fusion.
There are also methods that combine these two fusion manners, such as MVX-Net~\cite{sindagi2019mvx}.
Compared with proposal-level fusion, the point-level one is a more subtle manner, which is also adopted in our method for deep fusion.
Previous point-level fusion approaches~\cite{huang2020epnet,vora2020pointpainting} usually enhance the point feature from image semantics in a point-to-point manner, ignoring the surrounding context in 3D space.
Different from them, the proposed voxel field fusion represents the augmented image feature in a point-to-ray manner in the voxel field, which makes further use of merits from both modalities with sufficient context.

\section{Voxel Field Fusion}
\label{sec:method}
The overall framework is conceptually simple: {\em mixed augmentor} is designed to align data augmentation across modalities; 
{\em learnable sampler} is introduced to efficiently select key features for interaction;
and {\em ray-wise fusion} is proposed to fuse and combine features along the ray.

\subsection{Mixed Augmentor}
\label{sec:joint_aug}
Given inputs captured from the camera and LiDAR, we first process the data with correspondence, as presented in Figure~\ref{fig:main}.
To this end, a joint strategy, called mixed augmentor, is proposed to handle the aforementioned augmentation misalignment during training from two aspects, namely sample-added and sample-static augmentation.

\vspace{0.5em}
\noindent
\textbf{Sample-added.} 
The sample-added augmentation is defined to increase samples in each scene from the whole database, {\em i.e.,} GT-sampling~\cite{yan2018second}.
In this case, we supplement the RGB data of sampled 3D objects in a copy-paste manner~\cite{dvornik2018modeling,yun2019cutmix}.
That means for each sampled object, we crop data within the projected 2D box and paste it onto the input image, where the crops are reorganized according to the actual depth or the cropping order.
In this process, the occluded points covered by near samples are filtered to avoid the cross-modality ambiguity in nuScenes dataset, similar to that of~\cite{wang2021pointaugmenting}.

\vspace{0.5em}
\noindent
\textbf{Sample-static.} 
The sample-static augmentation includes a set of transformations without new sample added, {\em e.g.,} flipping, rescaling, and rotation.
Different from previous work~\cite{zhang2020moca,wang2021pointaugmenting}, which uses reprojection to find the pairwise correspondence across modalities, we utilize image-level operations for augmentations that affect pretrained 2D convolutions, as summarized in Table~\ref{tab:aug_type}.
Specifically, due to inherent properties like flipping and scaling variance of convolution, the asynchronous augmentation across modalities brings the misalignment.
For example, if a flipping operation is applied to point cloud $\mathcal{C}$ but not corresponding images $\mathcal{P}$, the left-right context of point $p_i$ projected from point $c_i$ would be malposed. 
We further validate effectiveness of the proposed workflow in Tables~\ref{tab:aug_comparison} and \ref{tab:augmentor}.

\begin{table}[t] 
\centering
\caption{Corresponding operations in the mixed augmentor.}
\begin{tabular}{lll}
  \toprule
  {\em type} & Point operation & Image operation\\ 
  \midrule
  Sample-added & GT-sampling & Copy-paste \\
  \midrule
  \multirow{3}{*}{Sample-static} & Flip & Image-flip \\
                               & Rescale & Image-rescale \\
                               & Rotate & Reproject \\
  
  \bottomrule
\end{tabular}
\label{tab:aug_type}
\end{table}

\subsection{Voxel Field Construction} 
\label{sec:constructor}
With the above designed augmentor, we have input image $\mathbf{X}^I\in\mathbb{R}^{H\times W\times 3}$ and voxelized point cloud $\mathbf{X}^V\in\mathbb{R}^{X\times Y\times Z\times 4}$, as depicted in Figure~\ref{fig:main}.
The feature encoder with stacked convolutions is applied to extract feature $\mathbf{F}_{l}^I$ and $\mathbf{F}_{l}^V$ for image and voxel in the $l$-th stage, where $l$ is set to 1 by default and further investigated in Table~\ref{tab:fusion_stage}.
With the ray constructor, the correspondence between voxel bin $v_i$ and image pixel $p_i$ is established in voxel field with the given projection matrix $\mathbf{T}_{\mathrm{Voxel\rightarrow{Image}}}$.

\vspace{0.5em}
\noindent
\textbf{Voxel Field.}
In voxel representation, point clouds of a scene are captured within a voxel space $\mathcal{V}$ that contains several bins. 
Every single voxel $v$ in it can be represented by the function $\mathcal{F}$, called voxel field.
Specifically, for the feature $\mathrm{F}^V_{l,v}$ in voxel bin $v$ with coordinates $(x,y,z)$, we have $\mathrm{F}^V_{l,v}=\mathcal{F}(x,y,z)$.
In the voxel field, the ray $\mathcal{R}_i \in \mathcal{V}$ from the point $p_i$ through the voxel space $\mathcal{V}$ in a fixed direction is constructed with
\begin{equation}\label{equ:ray_construct}
p_i = v_j \mathbf{T}^{T}_{\mathrm{Voxel\rightarrow{Image}}},\; \forall v_j\in \mathcal{R}_i.
\end{equation}
\begin{figure}[t!] 
\centering
\begin{subfigure}[t]{0.48\linewidth}
\centering
\includegraphics[width=0.98\linewidth]{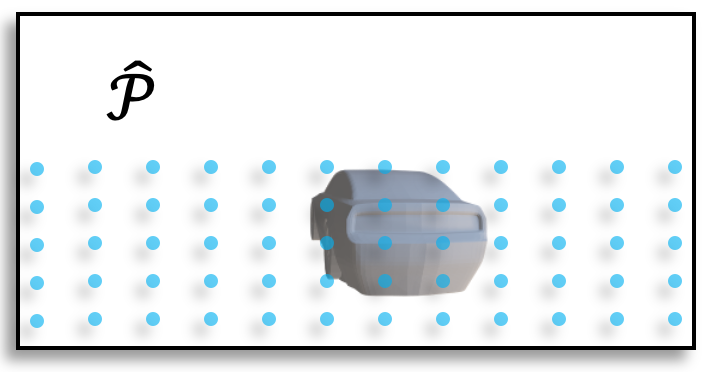}
\caption{Sample by uniformity}
\label{fig:sample_uniform}
\end{subfigure}
\hfill
\begin{subfigure}[t]{0.48\linewidth}
\centering
\includegraphics[width=0.98\linewidth]{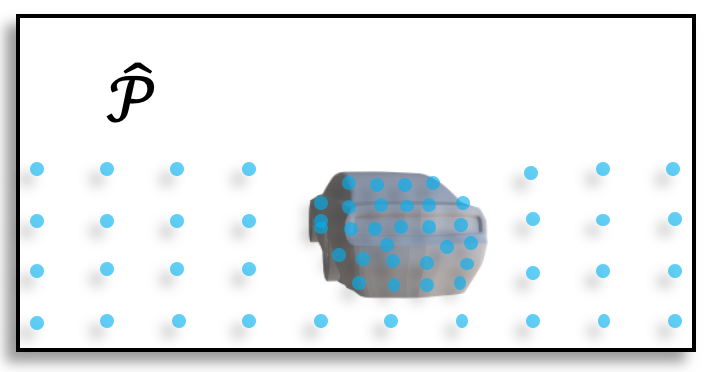}
\caption{Sample by density}
\label{fig:sample_density}
\end{subfigure}

\begin{subfigure}[t]{0.48\linewidth}
\centering
\includegraphics[width=0.98\linewidth]{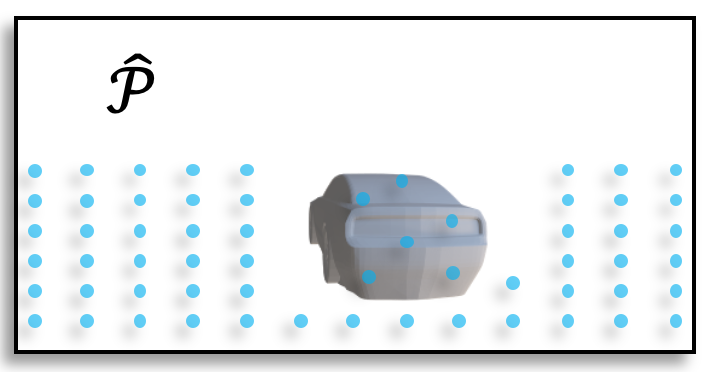}
\caption{Sample by sparsity}
\label{fig:sample_sparsity}
\end{subfigure}
\hfill
\begin{subfigure}[t]{0.48\linewidth}
\centering
\includegraphics[width=0.98\linewidth]{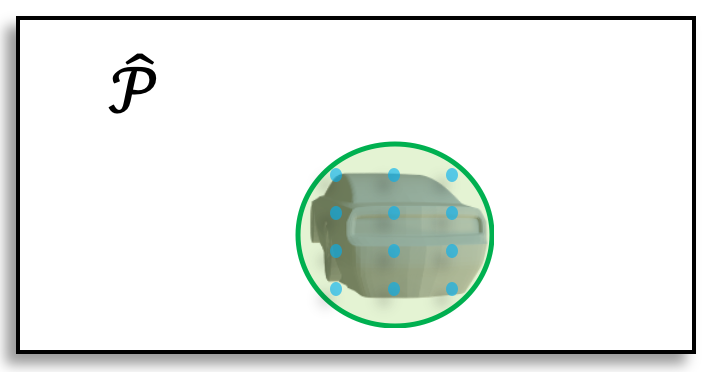}
\caption{Sample by importance}
\label{fig:sample_importance}
\end{subfigure}

\caption{Toy examples of different sampling methods. The blue points denote the sampled pixels for ray interaction. Compared with heuristic manners, our proposed learnable sampler in~\ref{fig:sample_importance} only considers the important green area with high responses.}
\label{fig:sample_method}
\end{figure}
This means all voxel bins $v_j$ with the same projected point $p_i$ are marked to be located in the $i$-th ray set $\mathcal{R}_i$.
Theoretically, the number of the whole set $\mathcal{R}$ can be up to $W\times H$ if without constraint, which brings huge computational cost that linearly increases with the number of sampled points.

\vspace{0.5em}
\noindent
\textbf{Learnable Sampler.}
To reduce the computational burden, learnable sampler is proposed to select $n$ points from the image plane for ray interaction, which brings $n$ rays in total.
In particular, we first split the image to several non-overlap windows with size $w\times w$ and filter out the empty windows with no projected point $p_i$, where $w$ is set to 64 by default.
Different from heuristic approaches for sampling, we adopt a learnable strategy to select key features by importance, as illustrated in Figure~\ref{fig:sample_method}.
For heuristic manner, $n$ features from the image plane are randomly sampled according to uniformity, density, and sparsity of projected LiDAR points.
Although the heuristic sampling method reduces the cost for ray construction, it still introduces several useless points that could increase computation in this process. 
To facilitate the efficiency, a learnable sampler $\mathcal{S}$ is further proposed, which {\em only} conducts the sampling procedure from the predicted important sub-region with high responses, as depicted in Figure~\ref{fig:sample_importance}.
Therefore, a set of sampled pixels $\widehat{\mathcal{P}}$ is achieved by 
\begin{equation}\label{equ:pixel_sample}
\mathcal{S}(\mathcal{P}) = \mathcal{U}(\{p_i: \mathbbm{1}(\delta(f(\mathrm{F}^I_{l,i})))=1\}),
\end{equation}
where $f$, $\delta$, and $\mathcal{U}$ denote stacked convolutions, $\mathrm{sigmoid}$ activation, and uniform sampler, respectively.
The indicator $\mathbbm{1}$ is set to 1 if the activated response in $p_i$ surpasses the threshold 0.5.
Considering the importance of foreground instance in 3D detection, here we set the Gaussian region inside each 2D object box to be the positive area for supervision, which is further explained in Section~\ref{sec:optimization}.
In this way, the amount of sampled pixels $\widehat{\mathcal{P}}$, as well as the cost, are further reduced.
Meanwhile, the high accuracy is still retained thanks to the proposed learnable sampler, as compared in Table~\ref{tab:sample_comparison}.

\begin{figure}[t!] 
\centering
\begin{subfigure}[t]{0.48\linewidth}
\centering
\includegraphics[width=0.98\linewidth]{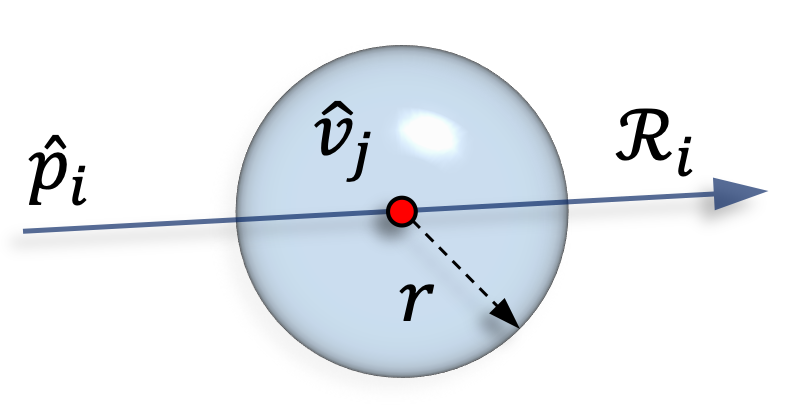}
\caption{Single fusion with each point}
\label{fig:fusion_single}
\end{subfigure}
\hfill
\begin{subfigure}[t]{0.48\linewidth}
\centering
\includegraphics[width=0.98\linewidth]{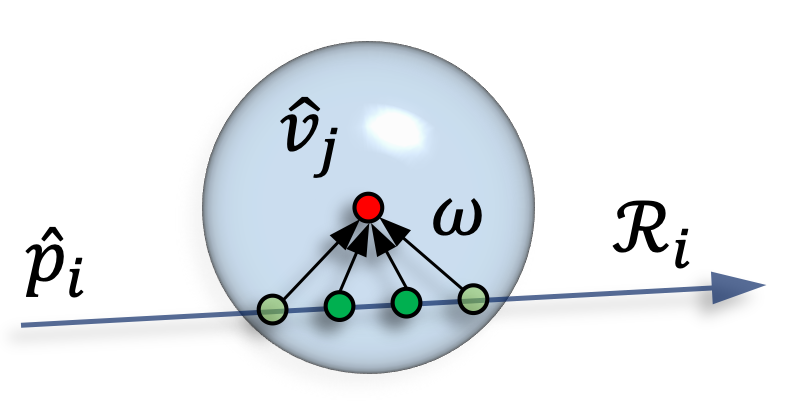}
\caption{Local fusion with aggregation}
\label{fig:fusion_aggregate}
\end{subfigure}

\begin{subfigure}[t]{0.48\linewidth}
\centering
\includegraphics[width=0.98\linewidth]{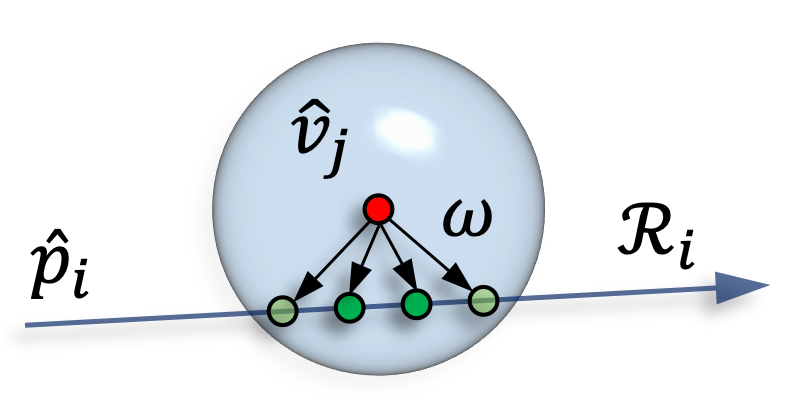}
\caption{Local fusion with propagation}
\label{fig:fusion_propagate}
\end{subfigure}
\hfill
\begin{subfigure}[t]{0.48\linewidth}
\centering
\includegraphics[width=0.98\linewidth]{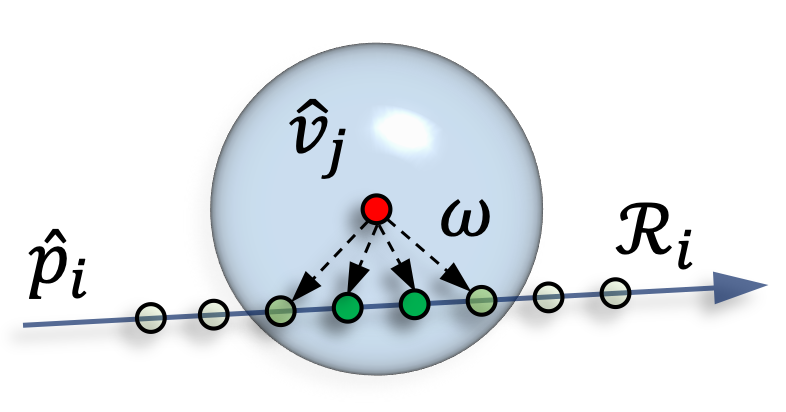}
\caption{Ray-wise fusion along the ray}
\label{fig:fusion_generate}
\end{subfigure}
\caption{Toy examples of different fusion methods. The red point $\hat{v}_j$ denotes the anchor voxel with LiDAR points. Green points in $\mathcal{R}_i$ indicate voxel bins $v_j$ within the ball with radius $r$. The dotted line in~\ref{fig:fusion_generate} represents the assigned existence probability for training.}
\label{fig:fusion_method}
\end{figure}
\subsection{Ray-voxel Interaction}
\label{sec:ray_fusion}
Recall from Equation~\eqref{equ:ray_construct}, the ray $\mathcal{R}_i$ is constructed with the pixel $\hat{p}_i\in \widehat{\mathcal{P}}$ from the above designed sampler $\mathcal{S}$.
Thus, the cross-modality fusion can be conducted with the ray $\mathcal{R}_i$ in the voxel field.
Previous research~\cite{liang2018continuous,vora2020pointpainting,wang2021pointaugmenting} performs fusion with sparse points from LiDAR sensor only, with no consideration of surrounding 3D context, as presented in Figure~\ref{fig:fusion_single}.
Therefore, a naive solution is to extend the perception region to include points that are located in the nearby region with radius $r$ of the voxel space $\mathcal{V}$, named local fusion.
In this section, we first introduce the basic local fusion and then improve over it to formulate the designed ray-wise fusion.

\vspace{0.5em}
\noindent
\textbf{Local Fusion.}
In this case, two types of local fusion are designed as our benchmark. 
Given the prior that features along the ray are more likely to be located near LiDAR points and the closer features usually contribute more, local fusion is conducted inside each Gaussian ball and ignores outside features along ray $\mathcal{R}_i$.
Here, we call the voxels that contains LiDAR points as {\em anchor voxels}.
The local fusion is divided into aggregation and propagation according to specific operations.
The aggregation manner in Figure~\ref{fig:fusion_aggregate} aggregates image features to the anchor voxel $\hat{v}_j$ with Gaussian weight $\omega$, while the propagation one in Figure~\ref{fig:fusion_propagate} propagates features in $\hat{v}_j$ to each voxel with weight $\omega$.

\vspace{0.5em}
\noindent
\textbf{Ray-wise Fusion.}
Although the designed local fusion extends the perception region from a single voxel to near area, it still set a hard boundary in this process. 
And such an approach cannot fully release the potential of ray-wise representation, especially in most of the voxels without LiDAR points.
Thus, besides the anchor voxel $\hat{v}_j$, we further extend the operation region to the {\em whole} ray, called {\em ray-wise fusion}.
Compared with single fusion in Figure~\ref{fig:fusion_single}, ours is more robust to sensor jitter due to a larger  fault-tolerant space brought by Gaussian ball.
Different from local fusion, the ray-wise manner in Figure~\ref{fig:fusion_generate} takes the above local prior to label assign in the training stage {\em only}, as elaborated in Section~\ref{sec:optimization}.
Specifically, given the voxel $v_j$ along the ray $\mathcal{R}_i$, its probability $\omega_j$ is calculated by
\begin{equation}\label{equ:predict_prob}
\omega_j = \delta(\left\langle\mathbf{F}^I_{l,i}, \mathbf{F}'_{l,v_j}\right\rangle),
\end{equation}
where $\delta$ denotes $\mathrm{sigmoid}$ activation, and the voxel feature $\mathbf{F}'_{l,v_j}=\mathrm{MLP}([x_j, y_j, z_j])$ is transformed from the coordinate $(x_j, y_j, z_j)$ of $v_j$ with multi-layer perceptron.
Here, $\omega_j$ can be viewed as the response of image feature $\mathbf{F}^I_{l,i}$ to the position of $v_j$.
From another perspective, this operation converts the monocular depth estimation with $\mathbf{F}^I_{l,i}$ from a regression problem to classification in a single ray $\mathcal{R}_i$. It actually reduces the solution space to a ray.
With the predicted score $\omega_j$, we conduct the fusion in voxel $v_j$ with 
\begin{equation}\label{equ:fusion}
\widehat{\mathcal{F}}(x_j, y_j, z_j) = \mathcal{F}(x_j, y_j, z_j) + \omega_j f([\mathbf{F}^I_{l,i}, \mathbf{F}'_{l,v_j}]).
\end{equation}
Here, $f$ denotes the convolution. $\widehat{\mathcal{F}}(x_j, y_j, z_j)$ indicates the new generated feature in voxel $v_j$. 
In this process, $\mathcal{F}(x_j, y_j, z_j)$ is set to 0 if the original voxel $v_j$ is empty, which can be viewed as completion in empty voxels of $\mathcal{V}$.
For network efficiency, only the voxels with top predicted scores $\omega$ are selected for fusion, which accounts for a quarter of original non-empty voxels in total.
As presented in Figure~\ref{fig:main}, the fused feature $\widehat{F}^V_l$ is obtained for the next stage in the 3D detection backbone and the following detection head.
This framework can be instantiated with various voxel-based backbones for experiments in Section~\ref{sec:experiment}, {\em e.g.,} PV-RCNN~\cite{shi2020pvrcnn}, Voxel R-CNN~\cite{deng2020voxelrcnn}, and CenterPoint~\cite{yin2021centerpoint}.

\subsection{Optimization Objectives}
\label{sec:optimization}
Recall from Equations~\eqref{equ:pixel_sample} and~\eqref{equ:predict_prob}, there are two learnable factors required  supervision.
For learnable sampler $\mathcal{S}$, given the prior that foreground objects are usually more important, we draw Gaussian distribution inside each box with
\begin{equation}\label{equ:gaussian_prob_2d}
\mathbf{Y}_{l,u,v} = \exp\left(-\frac{(u-\hat{u}_i)^2+(v-\hat{v}_i)^2}{2\sigma_i^2}\right),
\end{equation}
where $(\hat{u}_i, \hat{v}_i)$ denotes coordinate of the $i$-th object center, and $\sigma_i$ indicates the object size-adaptive standard deviation.
For activated probability $\omega_j$, the Gaussian-like supervision $\hat{\omega}_j$ of voxel $v_j$ is formulated as 
\begin{equation}\label{equ:3dgaussian_prob_3d}
\hat{\omega}_j = \exp\left(-\frac{(x-\hat{x}_j)^2+(y-\hat{y}_j)^2+(z-\hat{z}_j)^2}{2\sigma_j^2}\right),
\end{equation}
where $(\hat{x}_j, \hat{y}_j, \hat{z}_j)$ denotes the location of anchor voxel $\hat{v}_j$ in Figure~\ref{fig:fusion_method}, and $\sigma_j$ indicates the size-adaptive standard deviation.
And the voxels with Euclidean distance larger than radius $r$ are assigned as 0, which is investigated in Table~\ref{tab:radius_comparison}.
In this way, positions of LiDAR points can be utilized to provide supervision for feature selection in the ray.
Finally, the objective function of voxel field fusion is defined as
\begin{equation}\label{equ:final_loss}
\mathcal{L}_\mathrm{vff} = \lambda_{\mathrm{s}}\mathrm{BCE}(f(\mathrm{F}^I_{l}), \mathbf{Y}_{l}) + \lambda_{\mathrm{r}}\sum^{m}\mathrm{FL}(\omega,\hat{\omega})/m,
\end{equation}
where $f(\mathrm{F}^I_{l})$ represents the convolved feature in Equation~\eqref{equ:pixel_sample}, and $m$ denotes the size of ray set $\mathcal{R}$. $\mathrm{BCE}$ and $\mathrm{FL}$ indicate the binary cross-entropy loss and focal loss~\cite{lin2017focal}, respectively.
$\lambda_{\mathrm{s}}$ and $\lambda_{\mathrm{r}}$ denote the balanced loss factor for sampler and ray-wise fusion.
The optimization target for the whole network is the summation of raw detection loss $\mathcal{L}_\mathrm{det}$ and voxel field fusion loss $\mathcal{L}_\mathrm{vff}$.

\section{Experiments}
\label{sec:experiment}
In this section, the experimental setup is first introduced.
Then, we give the analysis of each component on KITTI {\em val} set with PV-RCNN~\cite{shi2020pvrcnn} as the backbone.
Comparisons with previous work on nuScenes~\cite{caesar2020nuscenes} and KITTI~\cite{geiger2012kitti} datasets are reported in the end.

\subsection{Experimental Setup}
\label{sec:exp_setup}
\noindent
\textbf{Dataset.}
KITTI dataset~\cite{geiger2012kitti} is a widely-adopted multi-modality benchmark for 3D object detection, which provides synced LiDAR points and front-view camera images.
It contains 7,481 samples for training and 7,518 samples for testing, where the training samples are usually split to {\em train} set with 3,712 samples and {\em val} set with 3,769 samples. 
nuScenes dataset~\cite{caesar2020nuscenes} is a large-scale benchmark for autonomous driving with 1,000 scenes, which are divided into 700, 150, 150 scenes in the {\em train} set, {\em val} set, and {\em test} set, respectively.
Here, we use the synced data with 10 object categories that are collected from a 32-beam LiDAR and six cameras in a 360-degree field of view.

\vspace{0.5em}
\noindent
\textbf{Implementation Details.}
In this work, three different backbones are adopted to validate the proposed framework, {\em i.e.,} PV-RCNN~\cite{shi2020pvrcnn} and Voxel R-CNN~\cite{deng2020voxelrcnn} on KITTI dataset, and CenterPoint~\cite{yin2021centerpoint} on nuScenes dataset.
We follow corresponding architecture and training settings in each network.
In the proposed VFF, three convolutions and MLPs are applied to the learnable sampler in Equation~\eqref{equ:pixel_sample} and feature transformation in Equation~\eqref{equ:predict_prob}, where individual MLP is used for each camera view. 
For optimization, $\lambda_{\mathrm{s}}$ and $\lambda_{\mathrm{r}}$ in Equation~\eqref{equ:final_loss} is set to 2 and 5 in all our experiments.
Besides the non-empty voxels, we select features along each ray with probability $\omega$ bigger than 0.05 in the inference stage.

\subsection{Component-wise Analysis}
\label{sec:abla_study}
\noindent
\textbf{Aligned Augmentation.}
As elaborated in Section~\ref{sec:joint_aug}, aligned data plays a vital role in cross-modality consistency.
In Table~\ref{tab:aug_comparison}, we compare the joint strategy for the sample-static augmentation listed in Table~\ref{tab:aug_type}, where PV-RCNN and basic single fusion are adopted.
As compared in Table~\ref{tab:aug_comparison}, the designed image-level transformation contributes significantly to feature-variant augmentations, namely Flip and Rescale.
If flip augmentation is adopted only, the performance gain is up to {\bf 1.65}\% AP on moderate cases.

\begin{table}[t]
 \caption{Comparisons on different augmentations on the KITTI {\em val} set. {\em aug type} and {\em align type} denote the adopted sample-static augmentation and type of cross-modality alignment, respectively. }
\centering
\begin{tabular}{lccccc}
  \toprule
  \multirow{2}{*}{\em aug type} & \multirow{2}{*}{\em align type} & & \multicolumn{3}{c}{AP$_{3D}$@Car-R40 (IoU=0.7)}\\ \cline{4-6}
  & & & Easy & {\bf Moderate} & Hard\\
  \midrule
  None & -- & & 88.69 & 81.82 & 79.91 \\
  \midrule
  \multirow{2}{*}{+ Flip} & reproject    & & 89.26 & 82.50 & 82.27 \\
                         & {\bf ours}  & & 91.34 & {\bf 84.15} & 82.48 \\
  \midrule
  \multirow{2}{*}{+ Rescale} & reproject   & & 91.78 & 84.25 & 82.58 \\
                            & {\bf ours}  & & 91.68 & {\bf 84.53} & 82.59 \\
  \midrule
  \multirow{2}{*}{+ Rotate} & reproject   & & 91.30 & 84.45 & 82.64 \\
                           & {\bf ours}  & & 91.43 & {\bf 84.57} & 82.61 \\
  \bottomrule
\end{tabular}
 \label{tab:aug_comparison}
\end{table}

\begin{table}[t]
 \caption{Comparisons on different strategies in mixed augmentor on the KITTI {\em val} set. {\em strategy} and {\em fusion} denote the strategy in mixed augmentor and the usage of voxel field fusion, respectively.}
\centering
\begin{tabular}{lccccc}
  \toprule
  \multirow{2}{*}{\em strategy} & \multirow{2}{*}{\em fusion} & & \multicolumn{3}{c}{AP$_{3D}$@Car-R40 (IoU=0.7)}\\ \cline{4-6}
  & & & Easy & {\bf Moderate} & Hard\\
  \midrule
  \multirow{2}{*}{Original} & \xmark & & 86.40 & 75.47 & 71.32 \\
                            & \cmark & & 87.08 & {\bf 76.25} & 72.03 \\
  \midrule
  \multirow{2}{*}{+ Added} & \xmark & & 87.77 & 79.19 & 76.71 \\
                             & \cmark & & 89.56 & {\bf 82.53} & 80.00 \\
  \midrule
  \multirow{2}{*}{+ Static} & \xmark & & 91.53 & 84.36 & 82.29 \\
                           & \cmark & & 92.31 & {\bf 85.51} & 82.92 \\
  \bottomrule
\end{tabular}
 \label{tab:augmentor}

\end{table}

\begin{table}[t]
 \caption{Comparisons on different sampling types on the KITTI {\em val} set. {\em sample type} and {\em learn} denote the adopted sampler and learnable manner for feature selection in Section~\ref{sec:constructor}, respectively. }
\centering
\begin{tabular}{lccccc}
  \toprule
  \multirow{2}{*}{\em sample type} & \multirow{2}{*}{\em learn} & & \multicolumn{3}{c}{AP$_{3D}$@Car-R40 (IoU=0.7)}\\ \cline{4-6}
  & & & Easy & {\bf Moderate} & Hard \\
  \midrule
  Uniformity & \xmark & & 88.69 & 81.90 & 79.82 \\
  Sparsity & \xmark & & 89.09 & 81.68 & 79.85 \\
  Density & \xmark & & 88.72 & 82.01 & 81.42 \\
  \midrule
  Importance & \cmark & & 89.11 & {\bf 82.10} & 79.98 \\
  \bottomrule
\end{tabular}
 \label{tab:sample_comparison}

\end{table}

\vspace{0.5em}
\noindent
\textbf{Mixed Augmentor.}
The proposed mixed augmentor in Section~\ref{sec:joint_aug} maintains consistency in augmentation from {\em sample-added} and {\em sample-static} strategies.
Here, we investigate the augmentor with VFF in Table~\ref{tab:augmentor}.
In the mixed augmentor, sample-added strategy yields notable gain with 6.28\% AP.
With the sample-static manner, the network with VFF is further improved to 85.51\% AP on moderate cases.
For clear comparisons, the following ablation studies are conducted with sample-added strategy {\em only} by default.

\vspace{0.5em}
\noindent
\textbf{Learnable Sampler.}
To facilitate voxel field construction, learnable sampler is proposed in Section~\ref{sec:constructor}.
In Table~\ref{tab:sample_comparison}, different sampling methods are compared, which are divided into heuristic and learnable sets.
As presented in Figure~\ref{fig:sample_method}, the heuristic sampler contains the type of uniformity, sparsity, and density.
And the network with heuristic sampler achieves peak performance with 82.01\%AP when sampling based on density.
As for the learnable sampler, the proposed sampling with importance in Figure~\ref{fig:sample_importance} attains a superior result than the heuristic one with 82.10\% AP.

\begin{table}[t]
 \caption{Comparisons on different fusion strategies on the KITTI {\em val} set. {\em range} and {\em operation} denote the fusion range and specific operation, respectively. Single fusion is a subset of the others.}
\centering
\begin{tabular}{lccccc}
  \toprule
  \multirow{2}{*}{\em range} & \multirow{2}{*}{\em operation} & & \multicolumn{3}{c}{AP$_{3D}$@Car-R40 (IoU=0.7)}\\ \cline{4-6}
  & & & Easy & {\bf Moderate} & Hard\\
  \midrule
  Single & -- & & 88.69 & 81.82 & 79.91 \\
  \midrule
  \multirow{2}{*}{Local} & aggregate & & 89.00 & 82.12 & 81.52 \\
                         & propagate & & 89.16 & 82.07 & 81.17 \\
  \midrule
  Ray-wise & generate & & 89.56 & {\bf 82.53} & 80.00 \\
  \bottomrule
\end{tabular}
 \label{tab:fuse_comparison}

\end{table}

\begin{table}[t]
 \caption{Comparisons on different types of supervision on the KITTI {\em val} set. {\em radius} and {\em gaussian} denote the adopted radius $r$ and the usage of Gaussian distribution in Section~\ref{sec:ray_fusion}, respectively.}
\centering
\begin{tabular}{cccccc}
  \toprule
  \multirow{2}{*}{\em radius} & \multirow{2}{*}{\em gaussian} & & \multicolumn{3}{c}{AP$_{3D}$@Car-R40 (IoU=0.7)}\\ \cline{4-6}
  & & & Easy & {\bf Moderate} & Hard\\
  \midrule
  0 & \xmark & & 89.11 & 82.10 & 79.98 \\
  \midrule
  1 & \cmark & & 89.56 & {\bf 82.53} & 80.00 \\
  2 & \cmark & & 88.33 & 81.75 & 79.54 \\
  \bottomrule
\end{tabular}
 \label{tab:radius_comparison}

\end{table}

\begin{table}[t]
 \caption{Comparisons on different fusion positions on the KITTI {\em val} set. {\em stage} and {\em fusion} denote the specific fusion stage in the network and the usage of voxel field fusion, respectively.}
\centering
\begin{tabular}{cccccc}
  \toprule
  \multirow{2}{*}{\em stage} & \multirow{2}{*}{\em fusion} & & \multicolumn{3}{c}{AP$_{3D}$@Car-R40 (IoU=0.7)}\\ \cline{4-6}
  & & & Easy & {\bf Moderate} & Hard\\
  \midrule
  -- & \xmark & & 87.77 & 79.19 & 76.71 \\
  \midrule
  Stage-1 & \cmark & & 89.56 & {\bf 82.53} & 80.00 \\
  Stage-2 & \cmark & & 89.12 & 80.43 & 79.70 \\
  Stage-3 & \cmark & & 89.64 & 80.27 & 77.93 \\
  Stage-4 & \cmark & & 88.69 & 80.22 & 78.10 \\
  \bottomrule
\end{tabular}
 \label{tab:fusion_stage}

\end{table}

\begin{table*}[th]
 \caption{Comparisons on different methods with a single model on the KITTI {\em val} set. * denotes our result from official source code.}
\centering
\resizebox{0.98\textwidth}{40mm}{
\begin{tabular}{lccccccccccccc}
  \toprule
   \multirow{2}{*}{\em method} & &
   
   \multicolumn{3}{c}{AP$_{3D}$@Car-R40 (IoU=0.7)} & & 
   \multicolumn{3}{c}{AP$_{3D}$@Car-R11 (IoU=0.7)} & &
   \multicolumn{3}{c}{AP$_{BEV}$@Car-R40 (IoU=0.7)} 
   \\ \cline{3-5} \cline{7-9} \cline{11-13}
  & & Easy & {\bf Moderate} & Hard & & Easy & {\bf Moderate} & Hard & & Easy & {\bf Moderate} & Hard\\
  \midrule
  \multicolumn{13}{c}{\small \em LiDAR-based}\\
  \midrule
  SECOND~\cite{yan2018second}  & & -- & -- & -- & & 87.43 & 76.48 & 69.10 & & -- & -- & --  \\
  PointRCNN~\cite{shi2019pointrcnn}  & & -- & -- & -- & & 88.88 & 78.63 & 77.38 & & -- & -- & -- \\
  STD~\cite{yang2019std}  & & -- & -- & -- & & 89.70 & 79.80 & 79.30 & & -- & -- & --  \\
  PV-RCNN~\cite{shi2020pvrcnn}  & & 92.57 & 84.83 & 82.69 & & 89.35 & 83.69 & 78.70 & & 95.76 & 91.11 & 88.93 \\
  Voxel R-CNN~\cite{deng2020voxelrcnn}  & & 92.38 & 85.29 & 82.86 & & 89.41 & 84.52 & 78.93 & & 95.52 & 91.25 & 88.99 \\
  \midrule
  \multicolumn{13}{c}{\small \em LiDAR+RGB}\\
  \midrule
  UberATG-MMF~\cite{liang2019multi}  & & -- & -- & -- & & 88.40 & 77.43 & 70.22 & & -- & -- & --  \\
  3D-CVF~\cite{yoo20203dcvf}  & & 89.67 & 79.88 & 78.47 & & -- & -- & -- & & -- & -- & -- \\
  EPNet~\cite{huang2020epnet} & & 92.28 & 82.59 & 80.14 & & -- & -- & -- & & 95.51 & 88.76 & 88.36  \\
  \midrule
  PV-RCNN*           & & 91.53 & 84.36 & 82.29 & & 88.95 & 83.51 & 78.72 & & 92.82 & 90.43 & 88.41  \\
  \multicolumn{1}{>{\columncolor{mygray}[6pt][419.5pt]}l}{+ {\bf VFF}} & & 92.31 & {\bf 85.51} & 82.92 & & 89.45 & {\bf 84.21} & 79.13 & & 95.43 & {\bf 91.40} & 90.66 \\
  \midrule
  Voxel R-CNN*  & & 92.27 & 84.88 & 82.50 & & 89.46 & 83.61 & 78.80 & & 95.51 & 91.13 & 88.85  \\
 \multicolumn{1}{>{\columncolor{mygray}[6pt][419.5pt]}l}{+ {\bf VFF}}  & & 92.47 & {\bf 85.65} & 83.38 & & 89.51 & {\bf 84.76} & 79.21 & & 95.65 & {\bf 91.75} & 91.39  \\
  \bottomrule
\end{tabular}
 \label{tab:kitti_val}
}
\end{table*}

\begin{table}[t]
 \caption{Comparisons on cross-modality fusion of different categories on the KITTI {\em val} set. We report results of car, pedestrian, and cyclist with IoU=0.7, 0.5, and 0.5, respectively.}
\centering
\begin{tabular}{lccccc}
  \toprule
  \multirow{2}{*}{\em category} & \multirow{2}{*}{\em fusion} & & \multicolumn{3}{c}{AP$_{3D}$-R40}\\ \cline{4-6}
  & & & Easy & {\bf Moderate} & Hard\\
  \midrule
  \multirow{2}{*}{Car} & \xmark & & 91.53 & 84.36 & 82.29 \\
                       & \cmark & & 92.31 & {\bf 85.51} & 82.92 \\
  \midrule
  \multirow{2}{*}{Pedestrian} & \xmark & & 66.04 & 59.19 & 54.15 \\
                              & \cmark & & 73.26 & {\bf 65.11} & 60.03\\
  \midrule
  \multirow{2}{*}{Cyclist} & \xmark & & 91.31 & 72.18 & 67.60 \\
                           & \cmark & & 89.40 & {\bf 73.12} & 69.86 \\

  \bottomrule
\end{tabular}
 \label{tab:different_class}

\end{table}

\begin{table}[t]
 \caption{Comparisons on different types of pretrained 2D backbone on the KITTI {\em val} set. {\em model} denotes the adopted pretrained model in feature encoder with corresponding task.}
\centering
\begin{tabular}{lccccc}
  \toprule
   \multirow{2}{*}{\em model} & \multicolumn{3}{c}{AP$_{3D}$@Car-R40 (IoU=0.7)}\\ \cline{2-4}
  & Easy & {\bf Moderate} & Hard\\
  \midrule
  {\em Cls}: ResNet~\cite{he2016resnet}            & 91.96 & 85.33 & 84.24 \\
  {\em Det}: Faster R-CNN~\cite{ren2016faster}      & 91.98 & 85.11 & 82.52 \\
  {\em Seg}: DeepLabV3~\cite{chen2017deeplabv3}   & 92.31 & {\bf 85.51} & 82.92  \\
  \bottomrule
\end{tabular}
 \label{tab:pretrain_comparison}
\end{table}

\begin{table}[t]
 \caption{Comparisons on cross-modality fusion of different data on the KITTI {\em val} set. {\em beam num} and {\em fusion} denote LiDAR beam number and the usage of voxel field fusion, respectively.}
\centering
\begin{tabular}{lccccc}
  \toprule
  \multirow{2}{*}{\em beam num} & \multirow{2}{*}{\em fusion} & & \multicolumn{3}{c}{AP$_{3D}$@Car-R40 (IoU=0.7)}\\ \cline{4-6}
  & & & Easy & {\bf Moderate} & Hard\\
  \midrule
  \multirow{2}{*}{Beam-64} & \xmark & & 91.53 & 84.36 & 82.29 \\
                           & \cmark & & 92.31 & {\bf 85.51} & 82.92 \\
  \midrule
  \multirow{2}{*}{Beam-32} & \xmark & & 91.14 & 79.51 & 76.54 \\
                           & \cmark & & 92.20 & {\bf 82.36} & 79.81 \\
  \bottomrule
\end{tabular}
 \label{tab:beam_comparison}

\end{table}

\vspace{0.5em}
\noindent
\textbf{Ray-voxel Interaction.}
The interaction between ray and voxel is a core operation of the proposed framework in Section~\ref{sec:ray_fusion}.
Here, we compare different fusion methods presented in Figure~\ref{fig:fusion_method}.
As shown in Table~\ref{tab:fuse_comparison}, the performance improves with fusion range increases from a single point to the whole ray.
Compared with single fusion and local fusion, the ray-wise strategy surpasses them with a significant gap, which proves the effectiveness of ray-wise fusion.

\vspace{0.5em}
\noindent
\textbf{Ray-wise Supervision.}
Considering noise from sensor jitter or other issues, the Gaussian-like assignment is proposed to provide supervision in Figure~\ref{fig:fusion_generate} and Equation~\eqref{equ:3dgaussian_prob_3d}.
Different supervision strategies are compared in Table~\ref{tab:radius_comparison}.
It is clear that the Gaussian distribution with radius 1 results in the best performance 82.53\% AP.
And extra region provides wrong guidance to the ray that goes through each Gaussian ball, which harms feature localization in each ray.

\vspace{0.5em}
\noindent
\textbf{Fusion Stage.}
We further investigate the fusion stage of the proposed VFF in Table~\ref{tab:fusion_stage}.
Compared with the baseline, the designed fusion yields superior performance in each stage.
The early-stage fusion contributes more, which surpasses the baseline with {\bf 3.34}\% AP.
Meanwhile, fusion in the later stage brings less gain, which could be attributed to the low resolution for key feature selection and insufficient fusion.

\vspace{0.5em}
\noindent
\textbf{Different Categories.}
The ray-wise representation of image features in VFF introduces sufficient context for ambiguous examples.
In Table~\ref{tab:different_class}, we report comparisons with VFF on various categories. 
It is clear that the performance of each class has been improved especially for {\em pedestrian}, which is up to nearly 6\% AP on cases of all difficulties.

\vspace{0.5em}
\noindent
\textbf{Pretrained Network.}
In Table~\ref{tab:pretrain_comparison}, we analyze the pretrained 2D backbone, which provides the feature $\mathbf{F}^I_l$ in the feature encoder of Figure~\ref{fig:main}.
Here, we adopt all the augmentations and ResNet-50 based models for different tasks in Table~\ref{tab:pretrain_comparison}, {\em i.e.,} classification, detection, and semantic segmentation.
Compared with other tasks, the network~\cite{chen2017deeplabv3} provides better features if pretrained with semantic setting.

\begin{table*}[t]
 \caption{Comparisons on different methods with single model on the nuScenes {\em test} set.}
\centering
\resizebox{0.98\textwidth}{30mm}{
\begin{tabular}{lccccccccccccc}
    \toprule
    {\em method}  & mAP & NDS & Car & Truck & Bus & Trailer & C.V. & Ped. & Motor. & Bicycle & T.C. & Barrier \\
    \midrule
    \multicolumn{13}{c}{\small \em LiDAR-based}\\
    \midrule
    PointPillars~\cite{lang2019pointpillars} & 30.5 & 45.3 & 68.4 & 23.0 & 28.2 & 23.4 & 4.1 & 59.7 & 27.4 & 1.1 & 30.8 & 38.9 \\
    3DSSD~\cite{yang20203dssd} & 42.6 & 56.4 & 81.2 & 47.2 & 61.4 & 30.5 & 12.6 & 70.2 & 36.0 & 8.6 & 31.1 & 47.9 \\
    CBGS~\cite{zhu2019cbgs} & 52.8 & 63.3 & 81.1 & 48.5 & 54.9 & 42.9 & 10.5 & 80.1 & 51.5 & 22.3 & 70.9 & 65.7 \\
    CenterPoint~\cite{yin2021centerpoint}  & 60.3 & 67.3 & 85.2 & 53.5 & 63.6 & 56.0 & 20.0 & 84.6 & 59.5 & 30.7 & 78.4 & 71.1 \\
    \midrule
    \multicolumn{13}{c}{\small \em LiDAR+RGB}\\
    \midrule
    PointPainting~\cite{vora2020pointpainting} & 46.4 & 58.1 & 77.9 & 35.8 & 36.2 & 37.3 & 15.8 & 73.3 & 41.5 & 24.1 & 62.4 & 60.2 \\
    FusionPainting~\cite{xu2021fusionpainting} & 66.3 & 70.4 & 86.3 & 58.5 & 66.8 & 59.4 & 27.7 & 87.5 & 71.2 & 51.7 & 84.2 & 70.2 \\
    MVP~\cite{yin2021multimodal} & 66.4 & 70.5 & 86.8 & 58.5 & 67.4 & 57.3 & 26.1 & 89.1 & 70.0 & 49.3 & 85.0 & 74.8 \\
    PointAugmenting~\cite{wang2021pointaugmenting} & 66.8 & 71.0  & 87.5 & 57.3 & 65.2 & 60.7 & 28.0 & 87.9 & 74.3 & 50.9 & 83.6 & 72.6 \\
    \midrule
    \multicolumn{1}{>{\columncolor{mygray}[6pt][417.5pt]}l}{{\bf VFF} + CenterPoint} & {\bf 68.4} & {\bf 72.4} & 86.8 & 58.1 & 70.2 & 61.0 & 32.1 & 87.1 & 78.5 & 52.9 & 83.8 & 73.9 \\
    \bottomrule
\end{tabular}
 \label{tab:nuscene_comparison}
}
\end{table*}

\begin{table}[t]
 \caption{Comparisons on different methods on KITTI {\em test} set.}
\centering
\begin{tabular}{lccccc}
  \toprule
   \multirow{2}{*}{\em method} & & \multicolumn{3}{c}{AP$_{3D}$@Car-R40 (IoU=0.7)}\\ \cline{3-5}
  & & Easy & {\bf Moderate} & Hard\\
  \midrule
  \multicolumn{5}{c}{\small \em LiDAR-based}\\
  \midrule
  PointPillars~\cite{lang2019pointpillars} & & 82.58 & 74.31 & 68.99 \\
  PointRCNN~\cite{shi2019pointrcnn} & & 86.96 & 75.64 & 70.70 \\
  Part-$A^2$~\cite{shi2020parta2} & & 87.81 & 78.49 & 73.51 \\
  STD~\cite{yang2019std} & & 87.95 & 79.71 & 75.09 \\
  SA-SSD~\cite{he2020sassd} & & 88.75 & 79.79 & 74.16 \\
  PV-RCNN~\cite{shi2020pvrcnn} & & 90.25 & 81.43 & 76.82 \\
  Voxel R-CNN~\cite{deng2020voxelrcnn} & & 90.90 & 81.62 & 77.06 \\
  \midrule
  \multicolumn{5}{c}{\small \em LiDAR+RGB}\\
  \midrule
  MV3D~\cite{chen2017mv3d} & & 74.97 & 63.63 & 54.00 \\
  F-PointNet~\cite{qi2018frustum} & & 82.19 & 69.79 & 60.59 \\
  AVOD~\cite{ku2018avod} & & 83.07 & 71.76 & 65.73 \\
  UberATG-MMF~\cite{liang2019multi} & & 88.40 & 77.43 & 70.22 \\
  EPNet~\cite{huang2020epnet} & & 89.81 & 79.28 & 74.59 \\
  3D-CVF~\cite{yoo20203dcvf} & & 89.20 & 80.05 & 73.11 \\
  \midrule
  \multicolumn{1}{>{\columncolor{mygray}[6pt][141pt]}l}{{\bf VFF} + PV-RCNN} & & 89.58 & 81.97 & 79.17 \\
  \multicolumn{1}{>{\columncolor{mygray}[6pt][141pt]}l}{{\bf VFF} + Voxel R-CNN} & & 89.50 & {\bf 82.09} & 79.29 \\
  \bottomrule
\end{tabular}
 \label{tab:kitti_test}
\end{table}

\vspace{0.5em}
\noindent
\textbf{Sparse LiDAR.}
To verify the effectiveness of VFF with different LiDAR sparsity, we downsample the LiDAR points on the KITTI dataset to 32-beam following~\cite{you2020pseudo++}.
As presented in Table~\ref{tab:beam_comparison}, the proposed VFF achieves significant gain with {\bf 2.85}\% AP over the baseline.
For {\em hard} cases, the gap is enlarged to {\bf 3.27}\% AP.
It could attribute to the replenishment of empty voxels that lack LiDAR points.

\subsection{Main Results}
\label{sec:main_result}
\noindent
\textbf{nuScenes.}
We further report results on the large scale nuScenes {\em test} set.
As shown in Table~\ref{tab:nuscene_comparison}, the proposed method surpasses all previous approaches with {\bf 68.4}\% mAP and {\bf 72.4}\% NDS.
Compared with our strong backbone CenterPoint~\cite{yin2021centerpoint}, the performance gain brought by VFF is up to {\bf 8.1}\% mAP and {\bf 5.1}\% NDS.
As for ambiguous classes like {\em motorcycle} and {\em bicycle}, the gain is even up to {\bf 19}\% AP.

\vspace{0.5em}
\noindent
\textbf{KITTI.}
In Table~\ref{tab:kitti_val}, we carry out experiments on KITTI {\em val} set.
Compared with the baseline, our proposed VFF achieves consistent gain on various evaluation metrics and attains 85.51\% and 85.65\% AP with PV-RCNN~\cite{shi2020pvrcnn} and Voxel R-CNN~\cite{deng2020voxelrcnn}, respectively.
The result on KITTI {\em test} set is reported on Table~\ref{tab:kitti_test}.
Compared with previous fusion-based methods, the proposed VFF pushes the top performance to 81.97\% AP and 82.09\% AP with PV-RCNN and Voxel R-CNN as the backbone, respectively.
Thanks to the designed fusion manner, our method outperforms all previous models on {\em hard} cases with {\bf 79.29}\% AP, which attains {\bf 2.2}\% AP improvement over the baseline.

\section{Conclusion}
\label{sec:conclusion}
We have presented the voxel field fusion, a conceptually simple yet effective framework for cross-modality fusion in 3D object detection.
The key difference from prior works lies in that we maintain modality consistency by representing and fusing augmented image features as a ray in the voxel field.
In particular, inconsistency in feature representation for multi-modalities is eliminated with learnable sampler and ray-wise fusion.
Meanwhile, the mixed augmentor is developed to bridge the gap in cross-modality data augmentation.
Experiments on KITTI and nuScenes dataset prove the effectiveness of the proposed framework, which achieves consistent gains in various benchmarks and surpasses previous fusion-based models on both datasets.

\section{Acknowledgments}
We thank Yilun Chen and Tao Hu for their suggestions. 
This work is supported by The National Key Research and Development Program of China (No. 2017YFA0700800) and Beijing Academy of Artificial Intelligence (BAAI).

{\small
\bibliographystyle{ieee_fullname}
\bibliography{egbib}
}

\clearpage

\appendix

\section{Experimental Details}
Here, we provide more details in the {\em network architecture} and {\em training process} of the proposed voxel field fusion.

\subsection{Network Architecture}
\noindent
\textbf{Voxel-based Network.}
In our experiments, the voxel-based network is adopted as the main framework.
In particular, we conduct experiments with different networks on corresponding official implementations, {\em i.e.}, OpenPCDet for PV-RCNN~[\textcolor{citecolor}{27}] and Voxel R-CNN~[\textcolor{citecolor}{8}] on the KITTI~[\textcolor{citecolor}{10}] dataset, and CenterPoint~[\textcolor{citecolor}{44}] on the nuScenes~[\textcolor{citecolor}{2}] dataset.
And all the settings are kept identical with the original version, except the experiment with beam-32 LiDAR in Table~\textcolor{red}{11}, where the {\em number of voxels} is reduced to a half.

\vspace{0.5em}
\noindent
\textbf{Image-based Network.}
The image-based feature encoder is required in the proposed cross-modality fusion, as depicted in Figure~\textcolor{red}{2}.
For different models in Table~\textcolor{red}{10}, only the basic backbone with ResNet-50 is utilized to provide the feature $\mathbf{F}^I_l$ in the $l$-th stage.
Feature encoders for different modalities in Figure~\textcolor{red}{2} are aligned to have a same block depth.
Specifically, in the $l$-th stage of the voxel-based network, we first extract the image feature $\mathbf{F}^I_l$ from the corresponding $l$-th stage of the pretrained model.
Then, the feature $\mathbf{F}^I_l$ is upsampled to have the same stride of that in voxel feature $\mathbf{F}^V_l$.
Therefore, the cross-modality correspondence can be established in the calibrator of Figure~\textcolor{red}{2}.

\subsection{Training Process}
\noindent
\textbf{Detailed Settings.}
As declared before, for the voxel-based network, we follow all settings in the corresponding framework for fair comparisons.
Meanwhile, for the image-based network, we detach the grad from the original backbone and update the parameters of adopted blocks using grad from the voxel-based network only.
Due to the variance in datasets, different strategies are adopted for network evaluation.
In particular, for the KITTI dataset, we follow common practice in~[\textcolor{citecolor}{27},\textcolor{citecolor}{8}] and report the top performance of {\em car} in last 10 epoches for stable comparisons.
For the nuScenes dataset, we report the performance of the final optimized model.

\vspace{0.5em}
\noindent
\textbf{Augmentation.}
In Section~\textcolor{red}{3.1} of the main paper, image-level transformation is proposed in the sample-static strategy of mixed augmentor.
Specifically, for image-level operations in Table~\textcolor{red}{1}, we first randomly sample 100 projected points from LiDAR and calculate the affine matrix according to selected transformed points.
Then, the image-level augmentation is conducted with the affine matrix.
Due to the asymmetrical arrangement of six cameras, image-level augmentations in the nuScene dataset can not be well-aligned.
Thus, we apply the image-level augmentations on KITTI and keep the reproject manner on nuScenes dataset.

\begin{table}[t!]
 \caption{Comparisons with different methods for context enlargement on the KITTI {\em val} set. {\em operation} and {\em 3D} denote the operation to enlarge context and the action in 3D voxel space, respectively.}
\centering
\begin{tabular}{lccccc}
  \toprule
  \multirow{2}{*}{\em operation} & \multirow{2}{*}{\em 3D} & & \multicolumn{3}{c}{AP$_{3D}$@Car-R40 (IoU=0.7)}\\ \cline{4-6}
  & & & Easy & {\bf Moderate} & Hard\\
  \midrule
  Single & \xmark & & 91.43 & 84.57 & 82.61 \\
  \midrule
  Conv & \xmark & & 91.93 & 84.89 & 82.69 \\
  Deform Conv & \xmark & & 92.10 & 84.91 & 82.98 \\
  Attention & \xmark & & 91.29 & 83.78 & 82.14 \\
  \midrule
  Ray-wise & \cmark & & 92.31 & {\bf 85.51} & 82.92\\
  \bottomrule
\end{tabular}
 \label{tab:context_comparison}

\end{table}

\section{Additional Analysis}
\noindent
\textbf{Captured Context.}
In Table~\ref{tab:context_comparison}, we investigate approaches to enlarge the captured context.
In particular, compared with single fusion in Figure~\textcolor{red}{4a}, two types of methods are compared by capturing rich context in the image plane and 3D voxel space.
As illustrated in Table~\ref{tab:context_comparison}, additional operations in the image plane bring limited gain. 
It is attributed to the damage of adjacency relation in the voxel space, as analysed before. 
And the ray-wise manner achieves noticeable gains with geometry priors for context casting.

\begin{table*}[t]
 \caption{Comparisons with different numbers of views on the nuScenes {\em val} set. * denotes our result from the official source code.}
\centering
\resizebox{\textwidth}{20mm}{
\begin{tabular}{lccccccccccccc}
    \toprule
    {\em method}  & mAP & NDS & Car & Truck & Bus & Trailer & C.V. & Ped. & Motor. & Bicycle & T.C. & Barrrier \\
    \midrule
    CenterPoint* & 58.9 & 66.4 & 85.4 & 56.9 & 69.7 & 36.0 & 17.5 & 85.1 & 59.7 & 40.8 & 70.0 & 67.6 \\
    \midrule
    + {\bf VFF}-1 view & 58.6 & 66.1 & 85.4 & 57.1 & 70.7 & 37.2 & 17.1 & 84.5 & 59.9 & 38.8 & 69.0 & 66.4 \\
    + {\bf VFF}-3 view & 60.1 & 67.1 & 85.8 & 59.1 & 71.4 & 39.4 & 18.3 & 85.7 & 62.5 & 40.8 & 72.0 & 66.1 \\
    + {\bf VFF}-6 view & 63.4 & 68.7 & 86.4 & 60.1 & 71.7 & 38.8 & 23.8 & 86.2 & 71.5 & 52.2 & 75.8 & 68.0 \\
    \midrule
    + Fade~[\textcolor{citecolor}{34}] & 65.2 & 70.1 & 87.2 & 60.9 & 72.1 & 39.2 & 25.0 & 87.0 & 73.2 & 61.1 & 77.0 & 69.7 \\
    + Flip Test & {\bf 66.3} & {\bf 71.2} & 87.6 & 62.7 & 72.8 & 40.5 & 26.5 & 87.5 & 75.0 & 62.7 & 77.8 & 70.0 \\
    \bottomrule
\end{tabular}
 \label{tab:nuscene_comparison}
}
\end{table*}

\begin{figure*}[t!] 
\centering
\begin{subfigure}[t]{0.245\linewidth}
\centering
\includegraphics[width=\linewidth]{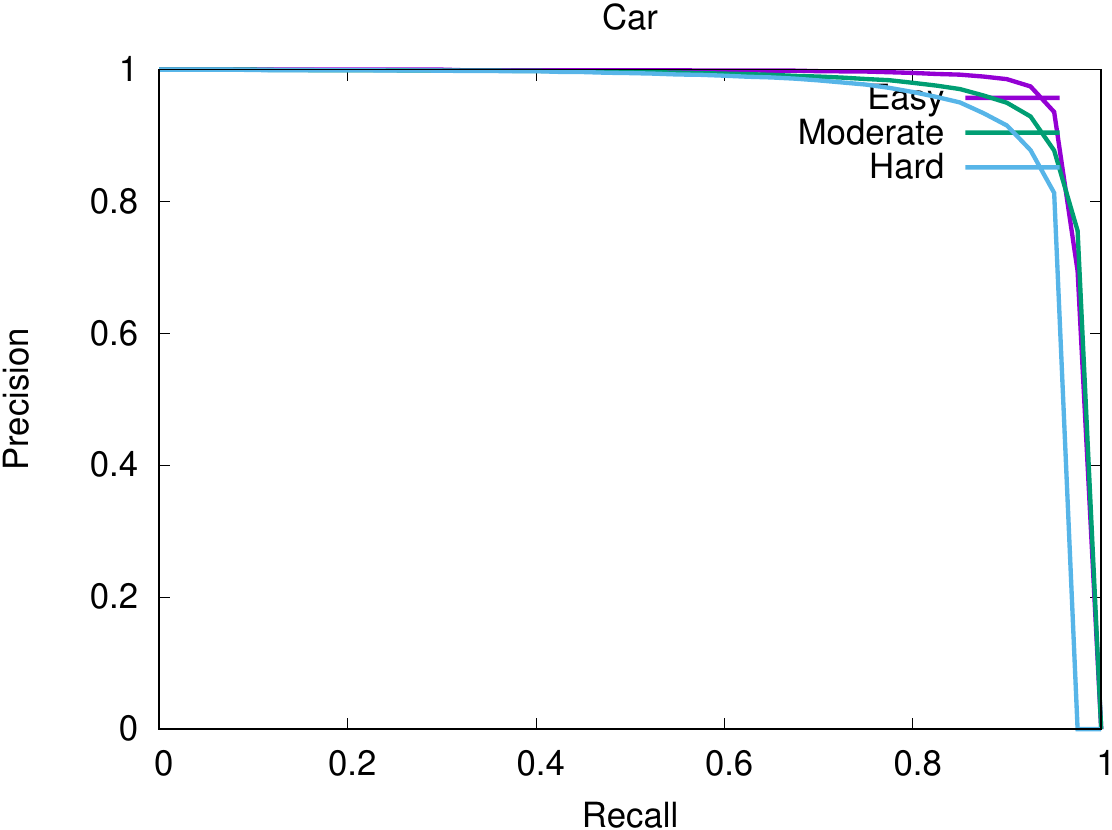}
\caption{Car: Detection}
\label{fig:appendix_car_det}
\end{subfigure}
\hfill
\begin{subfigure}[t]{0.245\linewidth}
\centering
\includegraphics[width=\linewidth]{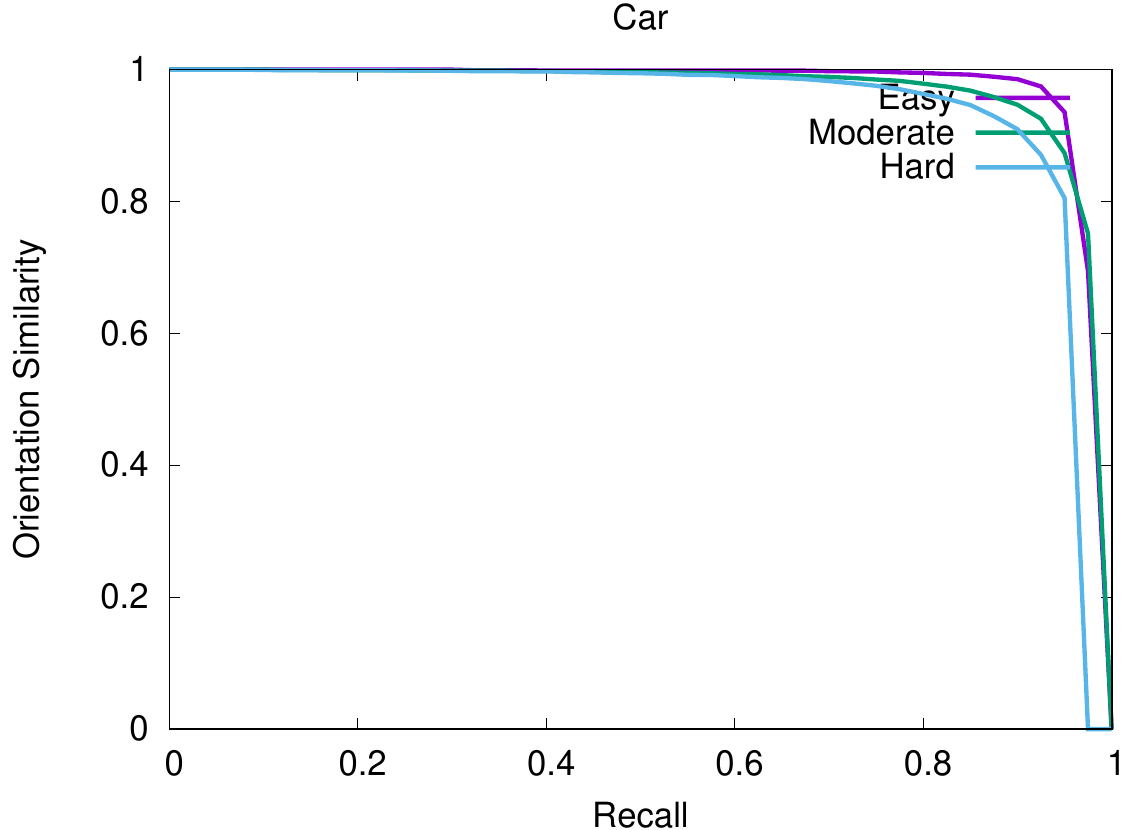}
\caption{Car: Orientation}
\label{fig:appendix_car_ori}
\end{subfigure}
\hfill
\begin{subfigure}[t]{0.245\linewidth}
\centering
\includegraphics[width=\linewidth]{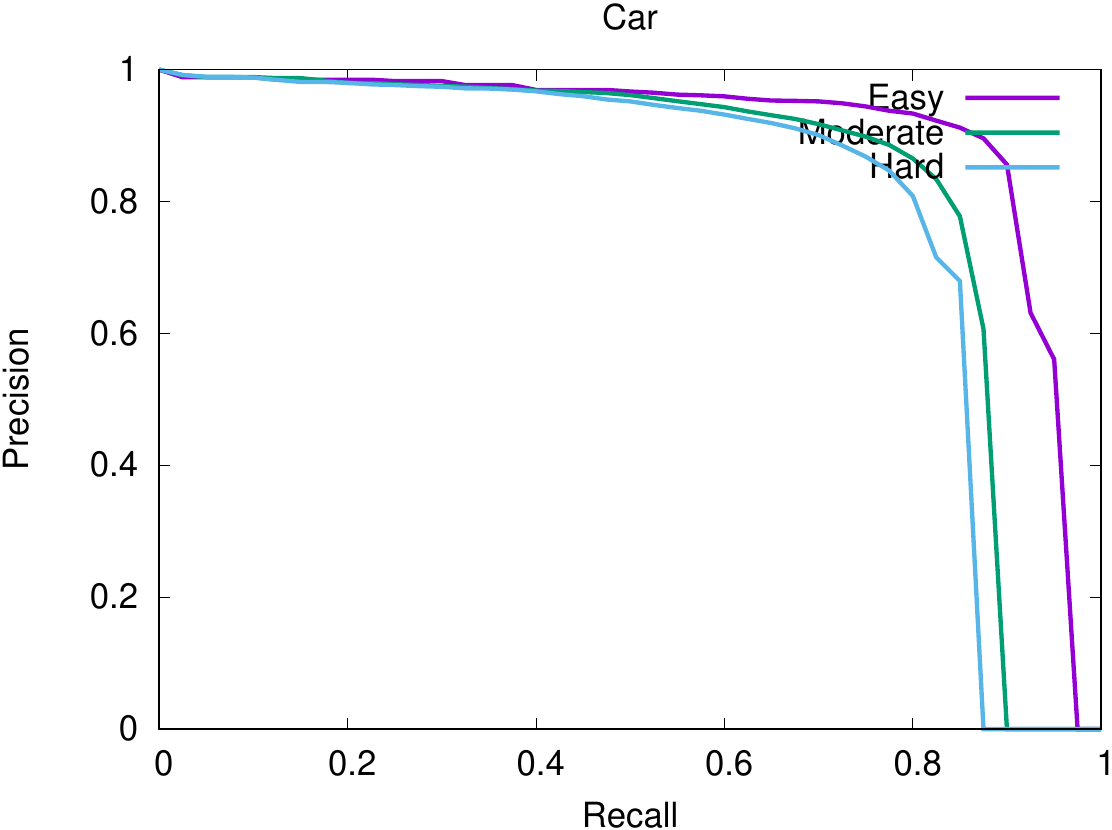}
\caption{Car: 3D Detection}
\label{fig:appendix_car_det3d}
\end{subfigure}
\hfill
\begin{subfigure}[t]{0.245\linewidth}
\centering
\includegraphics[width=\linewidth]{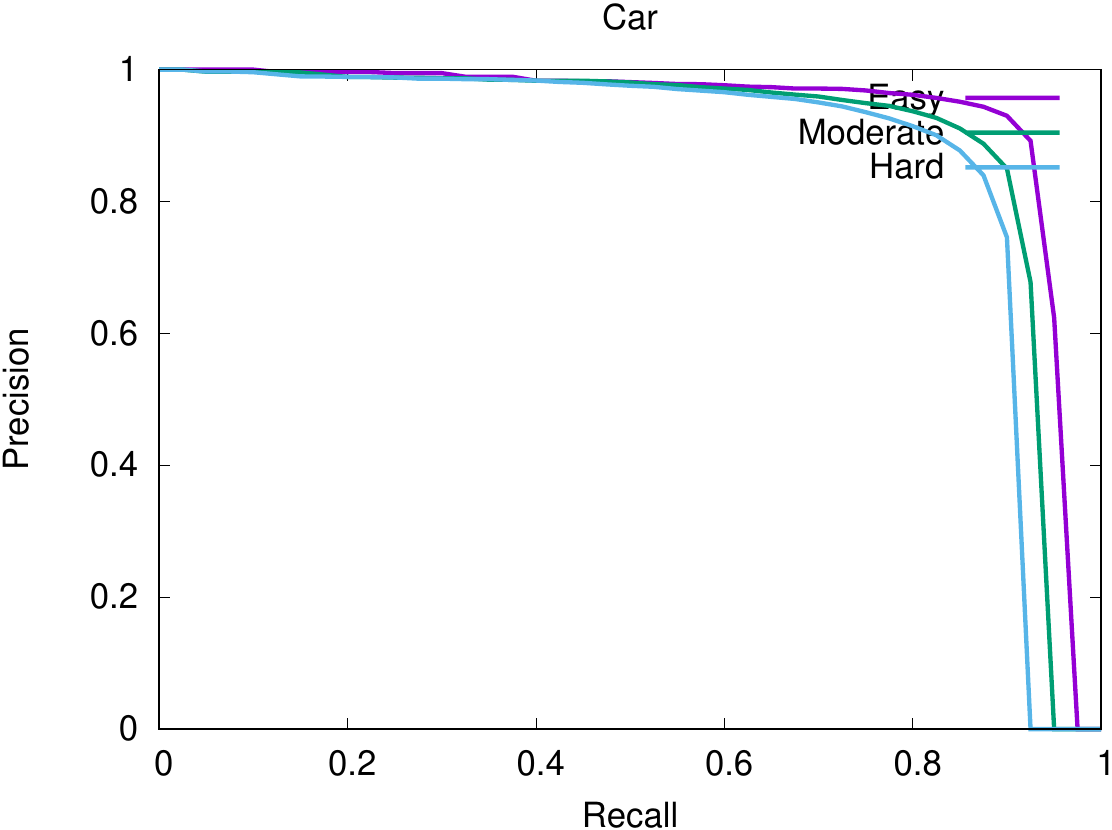}
\caption{Car: Bird's Eye View}
\label{fig:appendix_car_bev}
\end{subfigure}
\caption{Precision-Recall curves of Voxel R-CNN with the proposed voxel field fusion on the KITTI {\em test} set.}
\label{fig:appendix_kitti_result}
\end{figure*}

\begin{figure*}[t!] 
\centering
\begin{subfigure}[t]{0.245\linewidth}
\centering
\includegraphics[width=\linewidth]{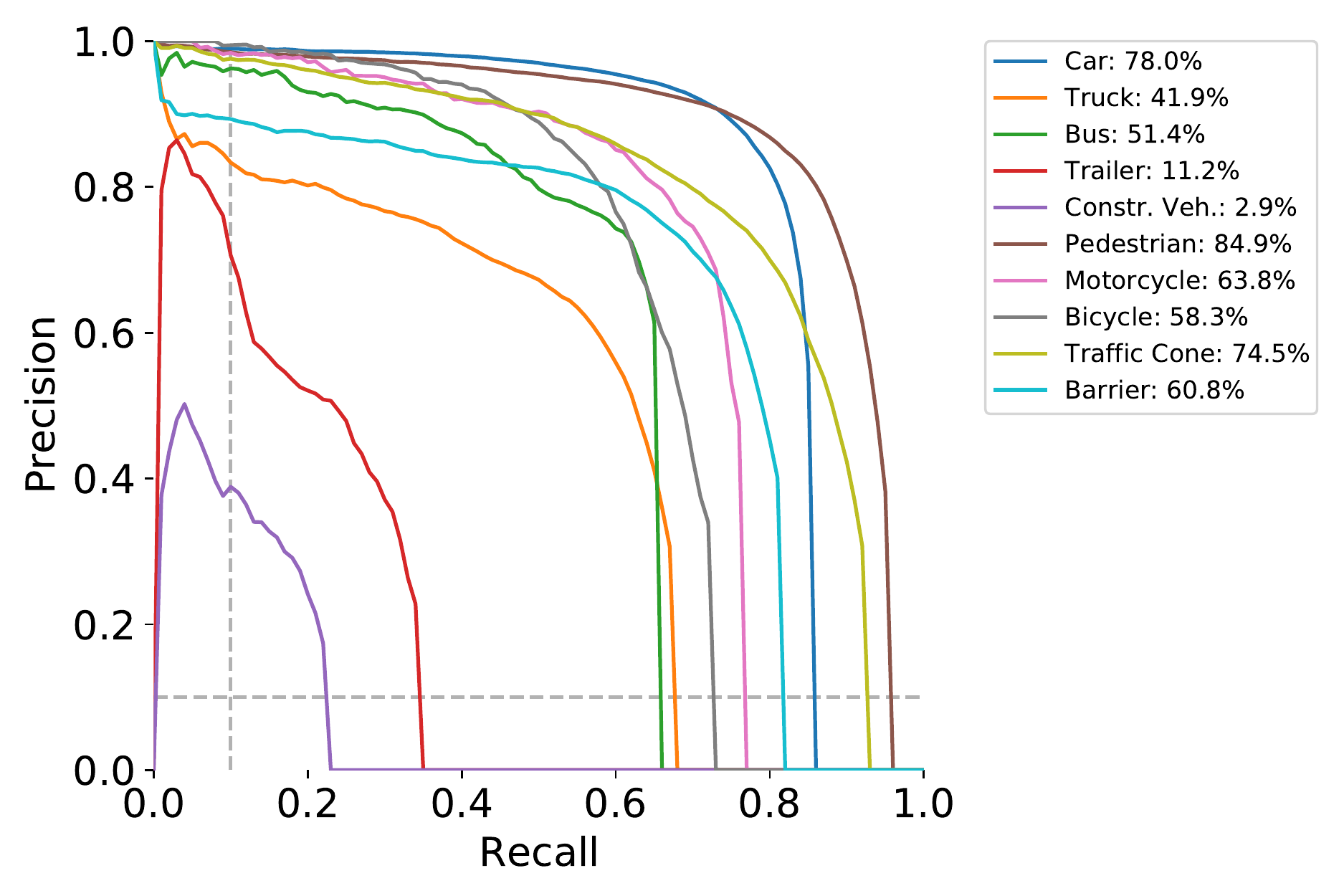}
\caption{Threshold: 0.5m}
\label{fig:appendix_dist_0_5}
\end{subfigure}
\hfill
\begin{subfigure}[t]{0.245\linewidth}
\centering
\includegraphics[width=\linewidth]{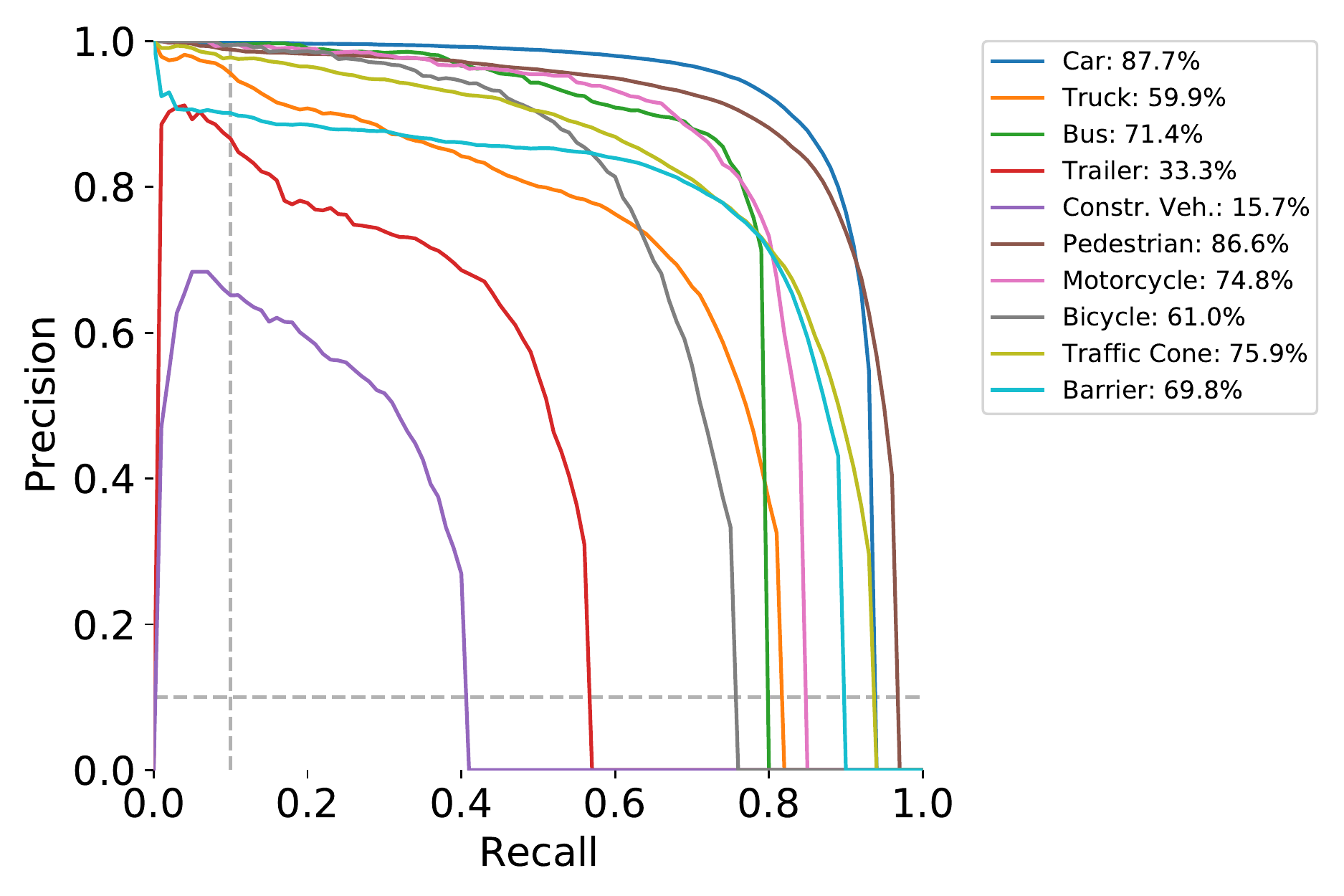}
\caption{Threshold: 1.0m}
\label{fig:appendix_dist_1_0}
\end{subfigure}
\hfill
\begin{subfigure}[t]{0.245\linewidth}
\centering
\includegraphics[width=\linewidth]{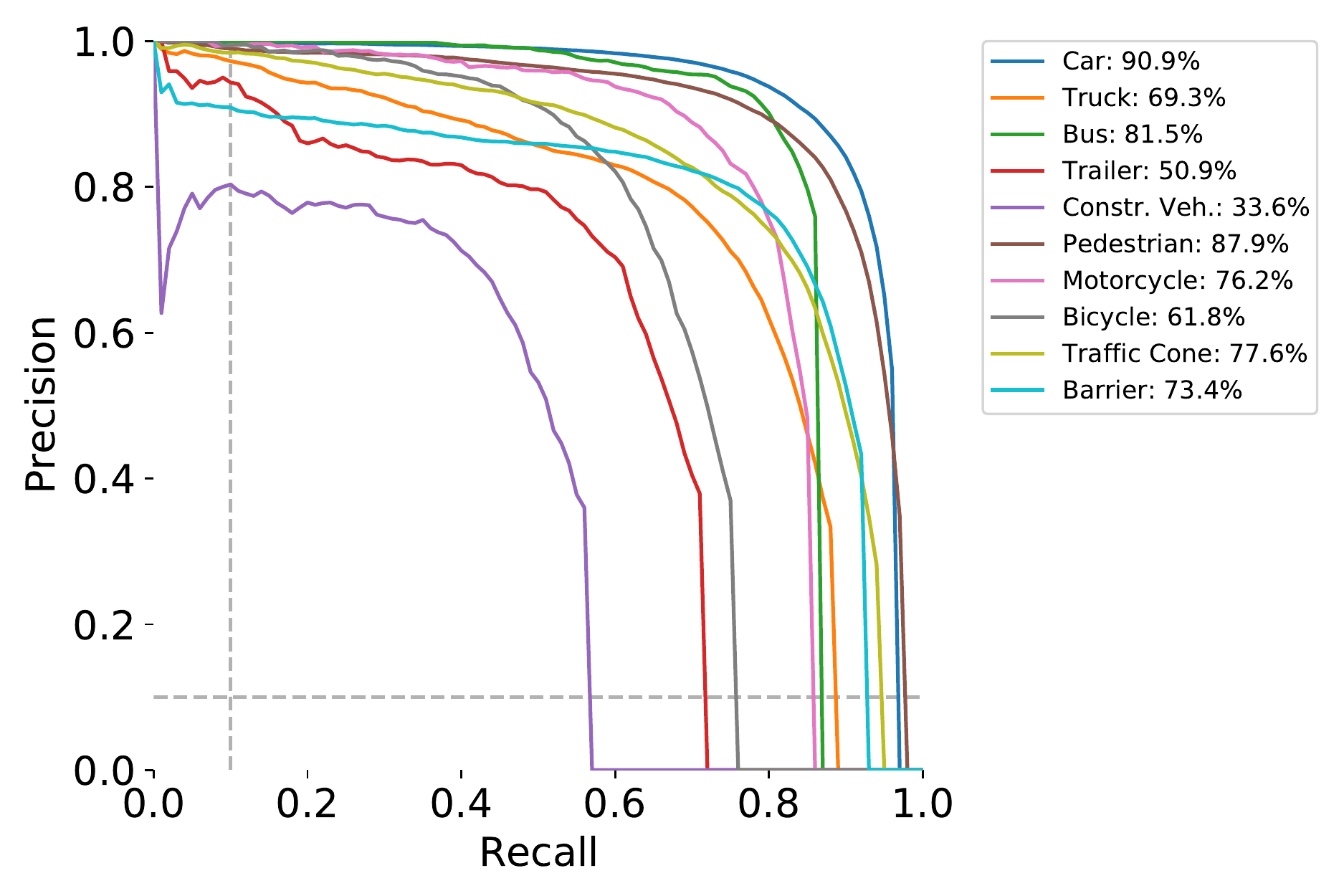}
\caption{Threshold: 2.0m}
\label{fig:appendix_dist_2_0}
\end{subfigure}
\hfill
\begin{subfigure}[t]{0.245\linewidth}
\centering
\includegraphics[width=\linewidth]{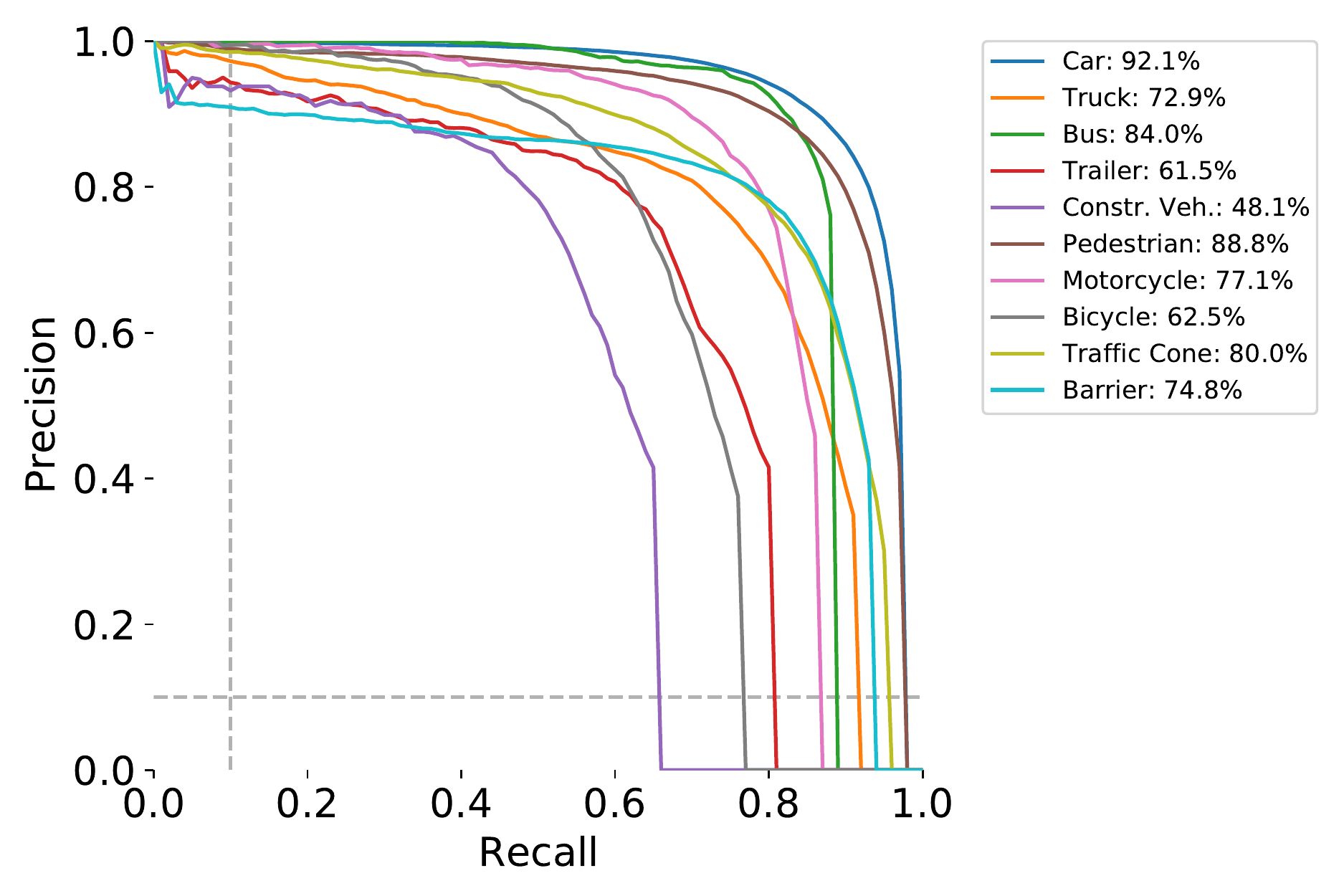}
\caption{Threshold: 4.0m}
\label{fig:appendix_dist_4_0}
\end{subfigure}
\caption{Precision-Recall curves of CenterPoint with the proposed voxel field fusion on the nuScenes {\em val} set.}
\label{fig:appendix_nuscenes_result}
\end{figure*}

\begin{figure*}[t!] 
\centering
\includegraphics[width=0.98\linewidth]{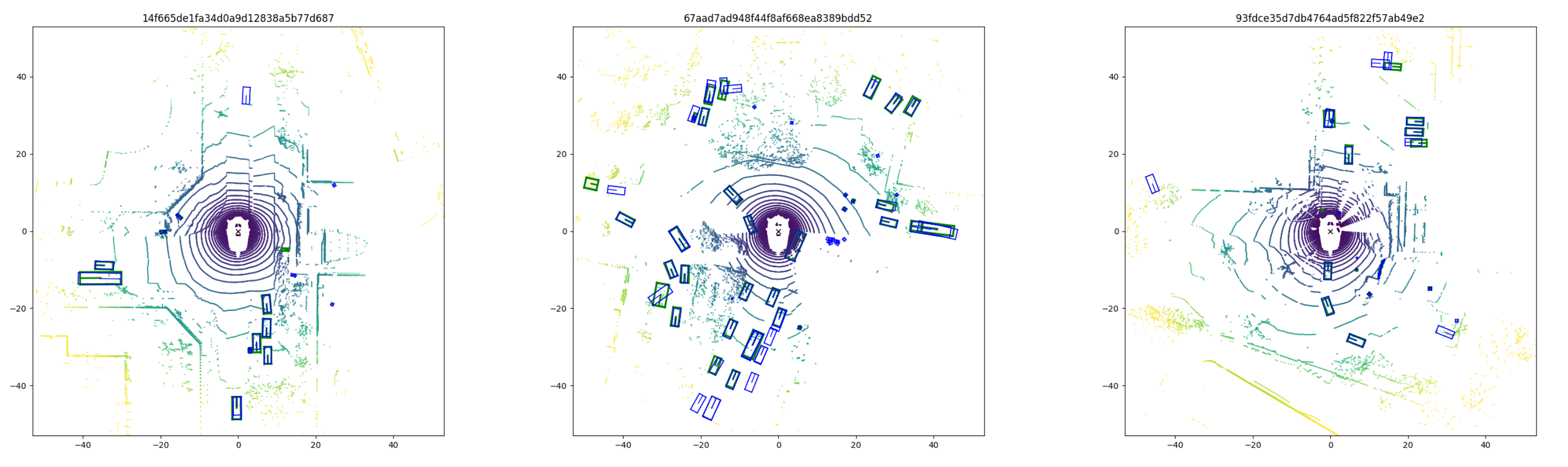}
\caption{Examples of the output predictions in different scenes on the nuScenes {\em val} set.}
\label{fig:appendix_vis_nuscenes}
\end{figure*}

\vspace{0.5em}
\noindent
\textbf{Number of Views.}
In Table~\ref{tab:nuscene_comparison}, we conduct experiments with different numbers of views on nuScenes dataset.
It is clear that the performance increases significantly with more views added, and attains top performance with {\bf 4.5}\% AP gain over the strong baseline.
An interesting finding is that fusion with one view brings little performance drop.
It could be attributed to the non-uniform distribution brought by part of image features in each scene.
With more training and testing strategies, the model achieves {\bf 66.3}\% mAP and {\bf 71.2}\% NDS on nuScenes {\em val} set.
Here, Fade strategy~[\textcolor{citecolor}{34}] denotes training {\em without} GT-Sampling in last 5 epoch.

\vspace{0.5em}
\noindent
\textbf{Detailed Results.}
In Figure~\ref{fig:appendix_kitti_result}, we draw precision-recall (PR) curves of car on KITTI {\em test} set.
The proposed method achieves satisfactory results on various metrics.
Meanwhile, we present PR curves with various thresholds on nuScenes {\em val} set.
It is obvious that the network attains good results for common vehicles like car and bus, while achieves inferior results in uncommon objects like construct vehicle.

\vspace{0.5em}
\noindent
\textbf{Predictions on nuScenes.}
As for large-scale scenes in the nuScenes~[\textcolor{citecolor}{2}] dataset, we present the predictions in Figure~\ref{fig:appendix_vis_nuscenes}.
It is clear that the proposed framework predicts satisfactory results with images from the 360-degree field of view.

\section{Qualitative Results}
In this section, qualitative studies of key components in the proposed framework are conducted with analysis.

\begin{figure*}[t!] 
\centering
\includegraphics[width=\linewidth]{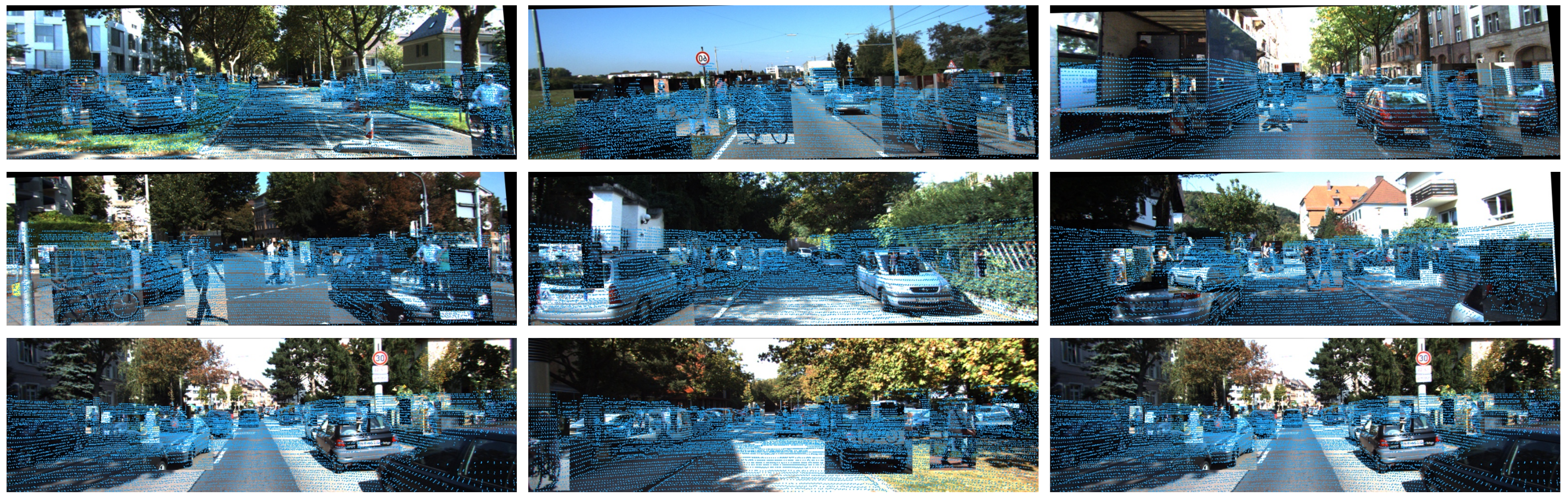}
\caption{Examples of the proposed mixed augmentor on KITTI {\em val} set. The blue points indicate the projected LiDAR in the image plane.}
\label{fig:appendix_aug}
\end{figure*}

\vspace{0.5em}
\noindent
\textbf{Mixed Augmentor.}
In Figure~\ref{fig:appendix_aug}, we present images with randomly adopted operations in the mixed augmentor.
It is clear that the cross-modality correspondence is well aligned in various scenes with designed image-level augmentations.

\begin{figure*}[t!] 
\centering
\begin{subfigure}[t]{\linewidth}
\centering
\includegraphics[width=\linewidth]{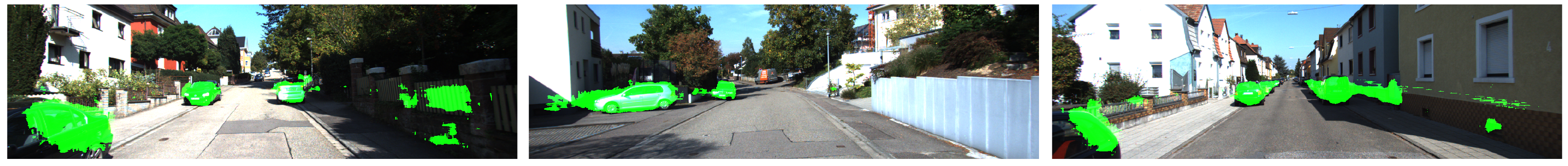}
\caption{Examples of important areas with the proposed learnable sampler. The green region in each figure indicates activated part with high responses.}
\label{fig:appendix_vis_kitti_0}
\end{subfigure}
\hfill
\begin{subfigure}[t]{\linewidth}
\centering
\includegraphics[width=\linewidth]{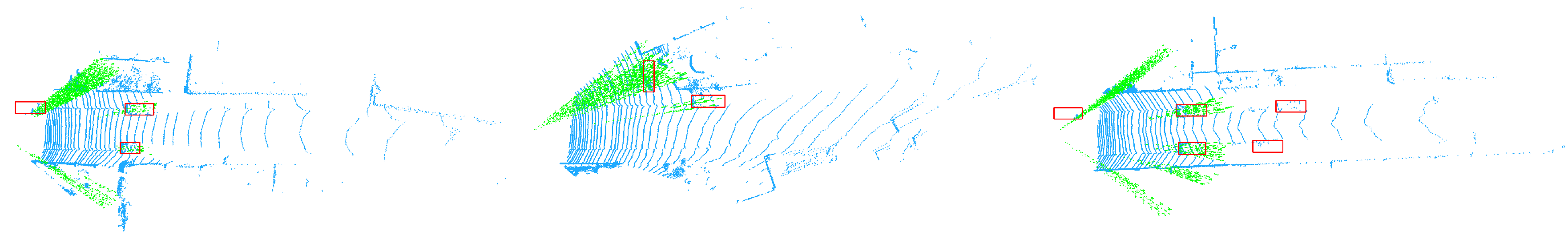}
\caption{Examples of selected points for ray-wise fusion in BEV. The green points locate newly generated features, and red boxes indicate annotated objects.}
\label{fig:appendix_vis_kitti_1}
\end{subfigure}
\hfill
\begin{subfigure}[t]{\linewidth}
\centering
\includegraphics[width=\linewidth]{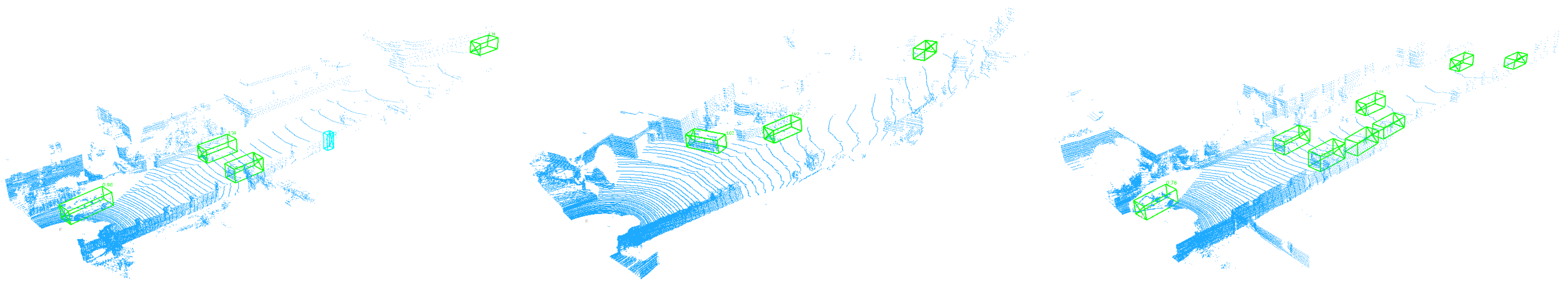}
\caption{Examples of the output predictions in different scenes. The green and light blue boxes represent the predicted cars and pedestrians.}
\label{fig:appendix_vis_kitti_2}
\end{subfigure}
\caption{Examples of key components in the voxel field fusion on KITTI {\em val} set. The figures in each column belong to the same scene.}
\label{fig:appendix_vis_kitti_result}
\end{figure*}

\vspace{0.5em}
\noindent
\textbf{Voxel Field Fusion.}
We further visualize predictions of each component, as shown in Figure~\ref{fig:appendix_vis_kitti_result}.
Given inputs with different modalities in various scenes, {\em learnable sampler} in Figure~\ref{fig:appendix_vis_kitti_0} and {\em ray-wise fusion} in Figure~\ref{fig:appendix_vis_kitti_1} works well.

\end{document}